\documentclass[10pt,journal,compsoc,twoside]{IEEEtran}
\ifCLASSOPTIONcompsoc
  \usepackage[nocompress]{cite}
\else
  \usepackage{cite}
\fi
\ifCLASSINFOpdf
\else
\fi
\usepackage{mathtools}
\usepackage{array}
\usepackage{float}
\usepackage{hyperref}
\usepackage{graphicx}
\graphicspath{{./graphics/}}
\usepackage{subfigure}
\usepackage{multicol}
\usepackage{multirow}
\usepackage{paralist}
\usepackage{bm}
\usepackage{amsmath,amsthm,amssymb,amsfonts}
\usepackage[usenames,dvipsnames]{color}
\usepackage{dsfont}
\usepackage{url}
\usepackage{rotating}
\usepackage{enumerate}
\usepackage{color}
\usepackage{slashbox}

\usepackage{ifpdf}
\usepackage{cite}

\usepackage{textcomp, booktabs}
\usepackage{colortbl}
\usepackage{gensymb}

\usepackage{times}
\usepackage{epsfig}
\usepackage{caption}
\usepackage{setspace}
\usepackage{threeparttable}

\makeatletter
\newif\if@restonecol
\makeatother

\usepackage[linesnumbered, ruled]{algorithm2e}%, vlined
\usepackage{algpseudocode}

\newtheorem{property}{Property}

\newenvironment{shrinkeq}[1]
{\bgroup
  \addtolength\abovedisplayshortskip{#1}
  \addtolength\abovedisplayskip{#1}
  \addtolength\belowdisplayshortskip{#1}
  \addtolength\belowdisplayskip{#1}}
{\egroup\ignorespacesafterend}

\newcommand{\bs}{\mathbf{s}}
\newcommand{\ba}{\mathbf{a}}
\newcommand{\be}{\mathbf{e}}

\begin{document}
\title{Self-Supervised Discovering of Interpretable Features for Reinforcement Learning}


\author{Wenjie~Shi,
        Gao~Huang,~\IEEEmembership{Member,~IEEE,}
        Shiji~Song,~\IEEEmembership{Senior~Member,~IEEE,}\protect\\
        Zhuoyuan~Wang,
        Tingyu~Lin,
        and~Cheng~Wu % <-this % stops a space
\IEEEcompsocitemizethanks{
\IEEEcompsocthanksitem W. Shi, G. Huang, S. Song, Z. Wang and C. Wu are with the Department of Automation, Tsinghua University, Beijing 100084, China.
E-mail: \{shiwj16, zhuoyuan16\}@mails.tsinghua.edu.cn;
        \{gaohuang, shijis, wuc\}@tsinghua.edu.cn.
(Corresponding author: Gao Huang.)
\IEEEcompsocthanksitem T. Lin is with the State Key Laboratory of Intelligent Manufacturing System Technology, Beijing Institute of Electronic System Engineering, Beijing 100039, China. E-mail: lintingyu2003@sina.com.
}
}


\markboth{IEEE TRANSACTIONS ON PATTERN ANALYSIS AND MACHINE INTELLIGENCE}%
{SHI \textit{et al.}: SELF-SUPERVISED DISCOVERING OF INTERPRETABLE FEATURES FOR REINFORCEMENT LEARNING}
\IEEEtitleabstractindextext{%
\begin{abstract}
 Deep reinforcement learning (RL) has recently led to many breakthroughs on a range of complex control tasks. However, the agent's decision-making process is generally not transparent. The lack of interpretability hinders the applicability of RL in safety-critical scenarios. While several methods have attempted to interpret vision-based RL, most come without detailed explanation for the agent's behavior. In this paper, we propose a self-supervised interpretable framework, which can discover interpretable features to enable easy understanding of RL agents even for non-experts. Specifically, a self-supervised interpretable network (SSINet) is employed to produce fine-grained attention masks for highlighting task-relevant information, which constitutes most evidence for the agent's decisions. We verify and evaluate our method on several Atari 2600 games as well as Duckietown, which is a challenging self-driving car simulator environment. The results show that our method renders empirical evidences about how the agent makes decisions and why the agent performs well or badly, especially when transferred to novel scenes. Overall, our method provides valuable insight into the internal decision-making process of vision-based RL. In addition, our method does not use any external labelled data, and thus demonstrates the possibility to learn high-quality mask through a self-supervised manner, which may shed light on new paradigms for label-free vision learning such as self-supervised segmentation and detection.
\end{abstract}

}

\maketitle

\IEEEdisplaynontitleabstractindextext

% \fi
\IEEEpeerreviewmaketitle

\IEEEraisesectionheading{\section{Introduction}\label{sec:introduction}}
 \IEEEPARstart{O}{ver} the last few years, deep reinforcement learning (RL) algorithms have achieved great success in a number of challenging domains, from video games \cite{mnih2015human, silver2016mastering, shi2019regularized} to robot navigation \cite{mirowski2016learning, shi2018multi, shi2018high}. In spite of their impressive performance across a wide variety of tasks, they are often criticized for being black boxes and lack of interpretability, which has increasingly been a pressing concern in deep RL. In addition, while deep RL substantially benefits from powerful function approximation capability of deep neural networks (DNNs), poor interpretability of which further exacerbates such concerns. Hence, developing the ability to understand the agent's underlying decision-making process is crucial before using deep RL to solve real-world problems where reliability and robustness are critical.

 In machine learning, there has been a lot of interest in explaining decisions of black-box systems \cite{cao2019interpretable, monfort2019moments, liu2019tabby, guidotti2019survey}. Some popular methods have provided visual explanations for DNNs, such as LIME \cite{ribeiro2016should}, LRP \cite{binder2016layer}, DeepLIFT \cite{shrikumar2016not},  Grad-CAM \cite{selvaraju2017grad}, Kernel-SHAP \cite{lundberg2017unified} and network dissection \cite{bau2017network, zhou2018interpreting}. However, these methods generally depend on class information and cannot be directly adapted to continuous RL tasks. In the context of vision-based RL, a feasible explanation approach is to learn t-Distributed Stochastic Neighbor Embedding (t-SNE) maps \cite{mnih2015human, zahavy2016graying, annasamy2019towards}, but which are difficult for non-experts to understand. Moreover, there are a number of works applying gradient-based \cite{wang2016dueling, zahavy2016graying} and perturbation-based \cite{greydanus2018visualizing} approaches to visualizing important features for RL agent's decisions, but the generated saliency maps are usually coarse and only provide limited quantitative evaluation. Another promising approach incorporates attention mechanisms into actor network to explain RL agent's decisions \cite{manchin2019reinforcement, mott2019towards, zhang2018agil, zhang2019atari}. However, these methods are not applicable to pretrained agent models whose internal structure cannot be changed anymore, additionally, some of these methods depend on human demonstration dataset \cite{zhang2018agil, zhang2019atari}.

 This paper aims to render reliable interpretation for vision-based RL where the agent's states are color images. To overcome the limitations of the above methods, we propose a self-supervised interpretable framework, which can discover interpretable features for easily understanding what information is task-relevant and where to look in the state. Answering these questions can provide valuable explanations about how decisions are made by the agent and why the agent performs well or badly. The main idea underlying our framework is novel and simple. Specifically, for a pretrained policy that needs to be explained, our framework learns to predict an attention mask to highlight the information (or features) that may be task-relevant in the state. If the generated actions are consistent when the policy takes as input the state and the attention-overlaid state respectively, the features highlighted by our framework are considered to be task-relevant and constitute most evidence for the agent's decisions.

 In this paper, the kernel module of our framework, i.e., a self-supervised interpretable network (SSINet), is first presented for vision-based RL agents based on two properties, namely \emph{maximum behavior resemblance} and \emph{minimum region retaining}. These two properties force the SSINet to provide believable and easy-to-understand explanations for humans. After the validity is empirically proved, the SSINet is applied to further explain agents from two facets including decision-making and performance. While the former focuses on explaining how the agent makes decisions, the latter emphasizes the explanations about why the agent performs well or badly. More concretely, the agent's decisions are explained by understanding basic attention patterns, identifying the relative importance of features and analyzing failure cases. Moreover, to explain the agent's performance, such as the robustness when transferred to novel scenes, two mask metrics are introduced to evaluate the generated masks, then how the agent's attention influences performance is explained quantitatively.

 We conduct comprehensive experiments on Atari 2600 video games \cite{bellemare2013arcade} as well as Duckietown \cite{gym_duckietown}, which is a challenging self-driving car simulator environment. Empirical results verify the effectiveness of our method, and demonstrate that the SSINet produces high-resolution and sharp attention masks to highlight task-relevant information that constitutes most evidence for the agent's behavior. In other words, our method discovers interpretable features for easily understanding how RL agents make decisions and why RL agents perform well or badly. Overall, our method takes a significant step towards interpretable RL.

 It is worth noting that our whole training procedure can be seen as self-supervised, because the data for training SSINet is collected by using the pretrained RL agent. Generally, self-supervised learning is challenging due to the lack of labelled data. It is not well understood why humans excel at self-supervised learning. For example, a child has never been supervised in pixel level, but can still perform highly precise segmentation tasks. Our method reveals a self-supervised manner to learn high-quality mask by directly interacting with the environment, which may shed light on new paradigms for label-free vision learning such as self-supervised segmentation and detection.

 The remainder of this paper is organised as follows. In the following two sections, we review related works and introduce the related background of RL. In Section \ref{sec:method}, we mainly present a self-supervised interpretable framework for vision-based RL. In Section \ref{sec:experiment}, empirical results are provided to verify the effectiveness of our method. In Section \ref{sec:applications} and \ref{sec:saliency metric}, our method is applied to explain how the agent makes decisions and why the agent performs well or badly, respectively. In the last section, we draw the conclusion and outline the future work.

%------------------------------------------------------------------------
\section{Related Work}\label{sec:related work}
\subsection{Explaining Traditional RL Agents}
 A number of prior works \cite{dodson2011natural, elizalde2008policy, hayes2017improving} have focused on explaining traditional RL agents. For example, based on the assumption that an exact Markov Decision Process (MDP) model is readily accessible, natural language and logic-based explanations are given for RL agents in \cite{dodson2011natural} and \cite{elizalde2008policy} respectively. More recently, execution traces \cite{hayes2017improving} of an agent are analyzed to extract explanations. However, these methods rely heavily on interpretable, high-level or hand-crafted state features, which is impractical for vision-based applications.

 Other explanation methods include decision tree \cite{gupta2015policy, bastani2018verifiable, roth2019conservative} and structural causal MDP \cite{waa2018contrastive, madumal2019explainable}. While decision tree can be represented graphically and thus aid in human understanding, a reasonably-sized tree with explainable attributes is difficult to construct, especially in the vision-based domain. Structural causal MDP methods are designed for specific MDP models and thus provide limited explanations.

\subsection{Explaining Vision-Based RL Agents}
 Explaining the decision-making process of RL agents has been a particular area of interest for recent works. Here we review prior works that aim to explain how inputs influence sequential decisions in vision-based RL. Broadly speaking, existing methods can be partitioned into four categories: embedding-based methods, gradient-based methods, perturbation-based methods and attention-based methods. In addition to those works that focus on the explanation of vision-based RL, some popular and relevant works for visual explanations of DNNs will also be reviewed.

\textbf{Embedding-based methods.}
 The main idea underlying embedding-based methods for interpreting vision-based RL is to visualize high dimensional data with t-SNE \cite{maaten2008visualizing}, which is a commonly used non-linear dimensionality reduction method. The simplest approach is to directly map the representation of perceptually similar states to nearby points \cite{mnih2015human, zahavy2016graying, annasamy2019towards}. Each state is represented as a point in the t-SNE map, and the color of the points is set manually using global features or specific hand crafted features. In addition, there is some work attempting to learn an embedded map where the distance between any two states is related to the transition probabilities between them \cite{engel2001learning}. However, an issue with these methods is that they emphasize t-SNE clusters or state transition statistics which are uninformative to those without a machine learning background.

\textbf{Gradient-based methods.}
 Methods in this category aim to understand what features of an input are most salient to its output by performing only one or a few backward passes through the network. The prototypical work is Jacobian saliency maps \cite{simonyan2014deep} where attributions are computed as the Jacobian with respect to the output of interest. Furthermore, there are several works generating Jacobian saliency maps and presenting it above the input state to understand which pixels in the state affect the value or action prediction the most \cite{wang2016dueling, zahavy2016graying}. Moreover, several other works modify gradients to obtain saliency maps for explanations of DNNs, such as Excitation Backpropagation \cite{zhang2018top}, Grad-CAM \cite{selvaraju2017grad}, LRP \cite{binder2016layer}, DeepLIFT \cite{shrikumar2016not} and SmoothGrad \cite{smilkov2017smoothgrad}. Unfortunately, Jacobian saliency maps may be difficult to interpret due to the change of manifold \cite{greydanus2018visualizing}, although they are efficient to compute and have clear semantics.

\textbf{Perturbation-based methods.}
 This category includes methods that measure the variation of a model's output when some of the input information is removed or perturbed \cite{zintgraf2017visualizing, fong2017interpretable}. The simplest perturbation approach computes attributions by replacing part of an input image with a gray square \cite{zeiler2014visualizing} or region \cite{ribeiro2016should}. An issue with this approach is that replacing pixels with a constant color introduces undesirable information. Another popular approach estimates feature importance by probing the model with randomly masked versions of the input image, such as RISE \cite{petsiuk2018rise}. A particular example of perturbation-based methods is Shapley values \cite{shapley1953value}, but the exact computation of which is NP-hard. Thus there are recent works applying perturbation approaches to approximating Shapley values for explaining DNNs, such as LIME \cite{ribeiro2016should}, Kernel-SHAP \cite{lundberg2017unified} and DASP \cite{ancona2019explaining}.

 In addition to these methods, there are some optimization-based perturbation methods \cite{fong2017interpretable, fong2019understanding, dabkowski2017real} aiming to optimize for the salient region using gradient descent technique. These methods are based on similar objective functions, but the specific implementations are different. The works \cite{fong2017interpretable, fong2019understanding} try to learn a mask to control the perturbation intensity of a Gaussian blur \cite{fong2017interpretable} for the input image. In fact, similar approach has been recently applied to visualize Atari agents by using masked interpolations between the original state and a Gaussian blur \cite{greydanus2018visualizing}. But a Gaussian blur perturbation fails to capture the shape of feature and results in coarse saliency maps. Another work \cite{dabkowski2017real} employs an encoder-decoder architecture for learning a mask to directly remove undesired regions, but it generally needs to use external labelled data and provides only limited quantitative evaluation. In this work, we build on these works and further propose a self-supervised framework to interpret RL agents both qualitatively and quantitatively, especially in terms of the underlying relationship between the agent's attention and the policy's performance, such as long-term return and generalization capability.

\textbf{Attention-based methods.}
 Another branch of that development is the incorporation of various attention mechanisms into vision-based RL agents. Learning attention to generate saliency maps for understanding internal decision pattern is one of the most popular methods \cite{wang2020paying} in deep learning community, and there are already a considerable number of works in the direction of interpretable RL. A mainstream approach is to augment the actor network (or agent) with customized self-attention modules \cite{sorokin2015deep, mousavi2016learning, yang2018learn, manchin2019reinforcement, nikulin2019free}, which learn to focus its attention on semantically relevant areas for making decisions. Another notable approach implements the key-value structure of attention to learn explainable policies by sequentially querying its view of the environment \cite{choi2017multi, annasamy2019towards, mott2019towards}. However, these methods generally need to change the agent's internal structure and thus cannot explain agent models that have already been trained. Moreover, there are some works attempting to build human Atari-playing attention dataset and use it to learn an explainable policy via imitation learning \cite{zhang2018agil, zhang2019atari}, but the cost can be prohibitive and it is impractical to do that for each vision-based RL task.

%------------------------------------------------------------------------
 \begin{table}[h]
  \centering
  % \fontsize{10.0}{11}\selectfont
  %\setlength{\tabcolsep}{1.5mm}{}
  \caption{Summary of Notation.}
  \label{tab:notation}
  \begin{threeparttable}
   \begin{tabular}{cl}
    \toprule
     Notation            & Description \cr
    \midrule
    $\bs_t$                & State at time step $t$   \cr
    $\tilde{\bs}_t$        & Attention-overlaid state at time step $t$  \cr
    $\ba_t$                & Action at time step $t$   \cr
    $R$                  & Return \cr
    $T$                  & Horizon \cr
    $\mathcal{S}$        & State space \cr
    $\mathcal{A}$        & Action space \cr
    $N_{\mathcal{A}}$    & Dimension of action space \cr
    $f_e$                & Encoder of SSINet or feature extractor \cr
    $f_d$                & Decoder of SSINet \cr
    $f_a$                & Action predictor of actor network \cr
    $f_\text{exp}$            & Explanation model or mask \cr
    $f^\theta_\text{exp}$     & Parameterized explanation model or mask \cr
    $\pi_e$              & Expert policy    \cr
    $\pi_m$              & Mask policy   \cr
    FOR                  & Feature Overlapping Rate   \cr
    BER                  & Background Elimination Rate   \cr
    $S_{e,f}$            & Area of extracted features   \cr
    $S_{t,f}$            & Area of ``true'' task-relevant features  \cr
    $S_{t,b}$            & Area of ``true'' background   \cr
    \bottomrule
   \end{tabular}
  \end{threeparttable}
 \end{table}
\section{Preliminaries}
 We consider a standard RL setup consisting of an agent interacting with an environment $E$ in discrete timesteps. Specifically, the agent takes an action $\ba_t$ in a state $\bs_t$, and receives a scalar reward $r_t$. Meanwhile, the environment changes its state to $\bs_{t+1}$. We model the RL task as a Markov decision process with state space $\mathcal{S}$, action space $\mathcal{A}$, initial state distribution $p(\bs_1)$, transition dynamics $p(\bs_{t+1}|\bs_t, \ba_t)$, and reward function $r(\bs_t, \ba_t)$. In all the tasks considered here the actions are real-valued $\ba_t\in \mathbb{R}^{N_\mathcal{A}}$.

 An agent's behaviour is defined by a policy $\pi$, which maps states to a probability distribution over the actions $\pi:~\mathcal{S}\rightarrow\mathcal{P}(\mathcal{A})$. The return from a state is defined as the sum of discounted future rewards, computed over a horizon $T$, i.e. $R_t=\sum_{i=t}^{T} \gamma^{i-t} r\left(\bs_i, \ba_i\right)$ with a discounting factor $\gamma\in[0,1]$. The return depends on the actions selected, and therefore on the policy $\pi$. The goal of an agent is to learn a policy $\pi$ which maximizes the following expected return from the start distribution
 \begin{shrinkeq}{-0.5ex}
  \begin{align}
   \label{eqn:RL objective} J = \mathbb{E}_{r_i,\bs_i\sim E,\ba_i\sim\pi}[R_0].
  \end{align}
 \end{shrinkeq}

 In this paper, we pretrain the agent with three model-free RL algorithms including proximal policy optimization (PPO) \cite{schulman2017proximal}, soft actor-critic (SAC) \cite{haarnoja2018soft} and twin delayed deep deterministic policy gradient (TD3) \cite{fujimoto2018addressing}. As an on-policy method, PPO uses trust region update to improve a general stochastic policy with gradient descent. Both SAC and TD3 are off-policy and based on actor-critic architecture. While SAC generally applies soft policy gradient \cite{shi2019soft} to learn a maximum entropy policy for capturing multiple modes of near-optimal behaviour, TD3 learns a deterministic policy by building on double Q-learning \cite{van2016deep} and deep deterministic policy gradient \cite{lillicrap2015continuous}.

%------------------------------------------------------------------------
 \begin{figure*}[t]
  \setlength{\abovecaptionskip}{-0.01cm}
  \setlength{\belowcaptionskip}{-0.25cm}
  \begin{center}
   \includegraphics[width=0.90\linewidth]{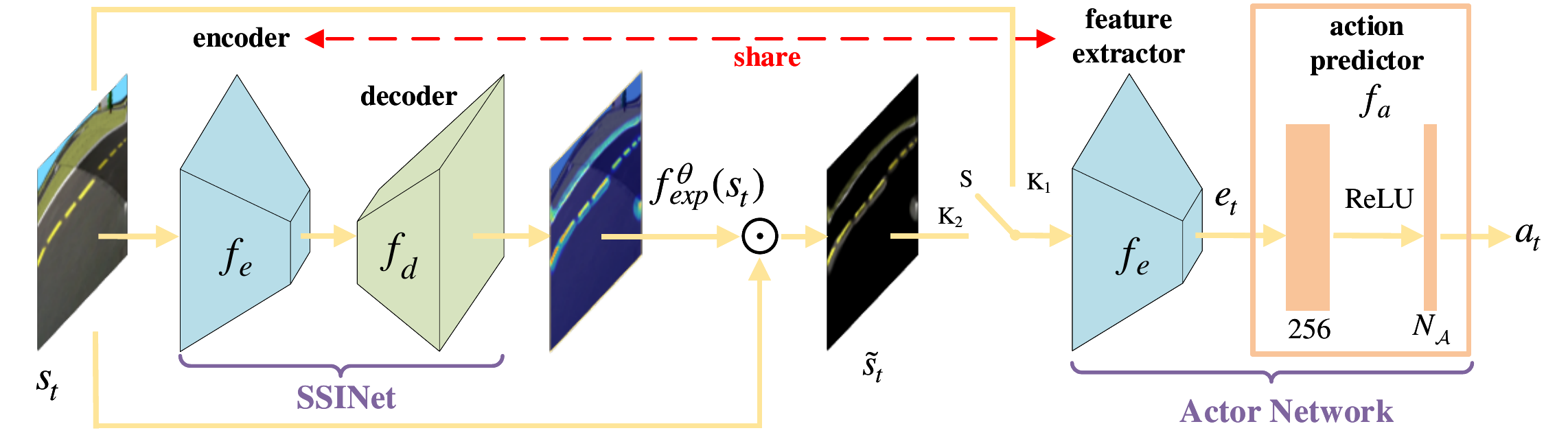}
  \end{center}
  \captionsetup{font={small}}
  \caption{Architecture diagram of our self-supervised interpretable framework, which includes two stages. In the first stage, we switch $S$ to $K_1$ and jointly pretrain the \emph{feature extractor} $f_e$ and \emph{action predictor} $f_a$ with RL. In the second stage, we switch $S$ to $K_2$ and train the \emph{decoder} $f_d$ while freezing $f_e$ and $f_a$. The \emph{encoder} is shared from the \emph{feature extractor} and also denoted by $f_e$. $N_{\mathcal{A}}$ is the dimension of action space.}
  \label{fig:sma2}
 \end{figure*}

%------------------------------------------------------------------------
\section{Method}\label{sec:method}
 In this section, we first present the main idea underlying our method for vision-based RL. Then, a self-supervised interpretable framework is proposed for a general RL agent. Finally, a two-stage training procedure is given for this framework. Table \ref{tab:notation} summarizes the main notations used in this work.

%------------------------------------------------------------------------
\subsection{Interpreting Vision-Based RL}\label{sec:interpretability}
 Consider a general setting where an expert policy is obtained by pretraining an actor (or agent) $\pi_e$, which takes as input an image $\bs_t$ to predict an action. Our goal is to train a separate explanation model $f_\text{exp}$ that can produce a mask $f_\text{exp}(\bs_t)$ corresponding to the importance assigned to each pixel of state $\bs_t$. In general, the mask $f_\text{exp}(\bs_t)$ can be explained as a kind of soft attention to show where the agent ``looks" to make its decision. In the context of vision-based RL, the explanation model $f_\text{exp}$ should satisfy two properties, namely \emph{maximum behavior resemblance} and \emph{minimum region retaining}.

\begin{property}[Maximum behavior resemblance]\label{pro:1}
 For an actor (or agent) $\pi_e$ and an explanation model $f_\text{exp}$, suppose $\tilde{\bs}_t$ is the attention-overlaid state corresponding to a specific state $\bs_t$, i.e., $\tilde{\bs}_t=f_\text{exp}(\bs_t)\odot \bs_t$, then
 \begin{shrinkeq}{-0.2ex}
  \begin{align}
   \label{eqn:property 1} \pi_e(\tilde{\bs}_t) \approx \pi_e(\bs_t) ~~(t=1,...,T),
  \end{align}
 \end{shrinkeq}
 where $\odot$ denotes the element-wise multiplication, and $\{\bs_1,...,\bs_T\}$ is an episode generated with $\pi_e$.
\end{property}

\begin{property}[Minimum region retaining]\label{pro:2}
 For a parameterized explanation model $f^\theta_\text{exp}$ and a specific state $\bs_t$, the retaining region is required to be minimum after overlaying the state $\bs_t$ with corresponding attention. That is
 \begin{shrinkeq}{-0.2ex}
  \begin{align}\label{eqn:property 2}
   \text{min}_{\theta}~\left\|f^\theta_\text{exp}(\bs_t)\right\|_1,
  \end{align}
 \end{shrinkeq}
 where $\|\cdot\|_1$ denotes the $L_1$-norm, and $\theta$ is the parameters of explanation model $f^\theta_\text{exp}$.
\end{property}

\noindent{\bf Remark}
 Property \ref{pro:1} requires the agent's behavior to be consistent with the original after the states are overlaid with the attentions generated by $f_\text{exp}$. Property \ref{pro:2} requires $f_\text{exp}$ to attend to as little information as possible, enabling easy understanding of decision-making for humans. In addition to the above properties, we emphasize that an explanation model for vision-based RL should be able to provide valuable explanations from two facets:

\noindent\textbf{Interpretability of decision-making.}
 In order for an agent to be interpretable, it must not only suggest informative explanations that make sense to  those without a machine learning background, but also ensure these explanations accurately represent the intrinsic reasons for the agent's decision-making. Concretely, it should be easy to understand how decisions are made, how an agent's current state affects its action, what information is used and where to look. While these questions are solved, the underlying decision-making process of RL agent is partially uncovered. Note that this type of analysis does not rely on the optimal policy; if an agent takes a suboptimal or even bad action, but the reasons for which can be explained faithfully, we still consider it interpretable.

\noindent\textbf{Interpretability of performance.}
 In the context of RL, transferability is whether the agent can generalize its policy across different scenes, and robustness is the ability of an agent to resist unknown external disturbances such as unexpected objects and new situations in novel scenes. In practice, it is meaningful and instructive to explain the performance of interest, especially when transferring the agent to novel scenes. More concretely, how the agent's attention influences performance. What factors will affect the performance? Do the RL algorithm and actor network architecture play a major role? Answering these questions can help explain why deep RL agents perform well or badly.

%------------------------------------------------------------------------
\subsection{Self-Supervised Interpretable Framework}\label{sec:sma2}
 In this section, we present a self-supervised interpretable framework for parameterized explanation model $f^\theta_\text{exp}$. As outlined in Fig. \ref{fig:sma2}, for an agent modelled by an actor network, we integrate a self-supervised interpretable network (SSINet) in front of the actor network. While the agent receives a state to predict an action at each time step, the SSINet produces an attention mask to highlight the task-relevant information for making decision without any external supervised signal. To that end, the SSINet must learn which parts of the state are considered important by the agent.

\textbf{SSINet.}
 Learning the mask is a dense prediction task, which arises in many vision problems, such as semantic segmentation \cite{ronneberger2015u, chen2017rethinking} and scene depth estimation \cite{mayer2016large}. Most of those approaches adopt an encoder-decoder structure. In order to make our masks sharp and precise, we build our SSINet by directly adapting a U-Net architecture \cite{ronneberger2015u} with only minor changes, exact details of which are described in Appendix \ref{app:network architecture} of the supplemental material. As depicted in Fig. \ref{fig:sma2}, our SSINet includes an \emph{encoder} $f_e$ and a \emph{decoder} $f_d$. Specifically, a state $\bs_t\in\mathbb{R}^{H\times W\times C}$ at time step $t$ (here a frame of height $H$, width $W$ and channel $C$) is encoded through $f_e$ to obtain low-resolution feature map, which is then taken as input of $f_d$ and upsampled to produce an attention mask $f^\theta_\text{exp}(\bs_t)\in[0, 1]^{H\times W\times 1}$:
 \begin{shrinkeq}{-0.2ex}
  \begin{align}
   \label{eqn:soft mask} f^\theta_\text{exp}(\bs_t) = \sigma\left(f_d(f_e(\bs_t))\right),
  \end{align}
 \end{shrinkeq}
 where $\sigma(\cdot)$ is the sigmoid nonlinearity. Afterwards, the attention mask $f^\theta_\text{exp}(\bs_t)$ is broadcast along the channel dimension of the state $\bs_t$, and element-wise multiplied with it to form a masked (or attention-overlaid) state $\tilde{\bs}_t$:
 \begin{shrinkeq}{-0.2ex}
  \begin{align}
   \label{eqn:state}     \tilde{\bs}_t = f^\theta_\text{exp}(\bs_t)\odot \bs_t,
  \end{align}
 \end{shrinkeq}
 where $\odot$ denotes the element-wise multiplication.

\textbf{Actor Network.}
 Generally, we use an actor network to model the policy of an RL agent. As shown in Fig. \ref{fig:sma2}, the actor network includes a \emph{feature extractor} $f_e$ and an \emph{action predictor} $f_a$. The \emph{feature extractor} $f_e$ takes as input a masked state $\tilde{\bs}_t$ (or a state $\bs_t$), and outputs low-resolution feature map $\be_t$. The \emph{action predictor} $f_a$ is a simple two-layer perception, which uses flatten feature map to predict an action $\ba_t\in\mathbb{R}^{\mathcal{N}_A}$. For clarify, we denote by \emph{expert policy} $\pi_e$ the actor network taking as input $\bs_t$, and denote by \emph{mask policy} $\pi_m$ the actor network taking as input $\tilde{\bs}_t$:
 \begin{shrinkeq}{-0.2ex}
  \begin{align}
   \label{eqn:expert policy}   \pi_e(\bs_t) &= \phi(f_a(f_e(\bs_t))), \\
   \label{eqn:mask policy}     \pi_m(\bs_t) &= \phi(f_a(f_e(\tilde{\bs}_t))),
  \end{align}
 \end{shrinkeq}
 where $\phi(\cdot)$ is the tanh nonlinearity for continuous RL tasks or the softmax nonlinearity for discrete RL tasks.

 There are several advantages of our self-supervised interpretable framework. First, our interpretable framework is applicable to any RL model taking as input visual images. Second, the SSINet learns to predict task-relevant information, which can provide intuitive and valuable explanations for the agent's decisions. Finally, we emphasize that the SSINet is a flexible explanatory module which can be adapted to other vision-based decision-making systems.

%------------------------------------------------------------------------
 \begin{algorithm}[t]
  \caption{Two-stage training procedure for SSINet}
  \label{alg:sma2}
  \SetAlgoNoLine
  Initialize $f_e$, $f_d$ and $f_a$\;
  Switch $S$ to $K_1$, obtain an expert policy $\pi_e$ by jointly pretraining $f_e$ and $f_a$\ with RL\;
  Generate state-action pairs $\{(\bs_i, \ba^l_i)\}^M_{i=1}$ using $\pi_e$\;
  Switch $S$ to $K_2$, freeze $f_e$ and $f_a$\;
  \For{each iteration}
  {
   Calculate the mask loss $L_\text{mask}$ (\ref{eqn:mask loss}) with collected state-action pairs\;
   Update $f_d$ using $L_\text{mask}$\;
  }
 \end{algorithm}
\subsection{Training Procedure}\label{sec:training}
 Our training procedure includes two stages. The first stage aims to obtain an expert policy $\pi_e$ and then use it to generate state-action pairs, which is used for training SSINet in the second stage. The objective of second stage is to learn interpretable attention masks $f^\theta_\text{exp}$ for explaining the agent's behaviour. Instead of learning $f^\theta_\text{exp}$ from scratch, we reuse the \emph{feature extractor} $f_e$ of actor network as the encoder of SSINet. This offers at least two advantages: (i) it speeds up the training since internal features from the RL agent can be reused; (ii) it greatly reduces the risk of overfitting and ensures that the generated mask is compatible with the RL agent to be explained. Overall, the whole training procedure is self-supervised, since there is no external labelled data.

 In the first stage, we switch $S$ to $K_1$ (as shown in Fig. \ref{fig:sma2}) and pretrain the \emph{feature extractor} $f_e$ and \emph{action predictor} $f_a$ with PPO \cite{schulman2017proximal} algorithm. After training, the resulting expert policy $\pi_e$ is used to collect data by generating $M$ state-action pairs $\{(\bs_i, \ba^l_i)\}^M_{i=1}$ with the action $\ba^l_i=\pi_e(\bs_i)$.

 In the second stage, we switch $S$ to $K_2$ and train the SSINet. The weights of the encoder is fixed during the second stage. Based on Property \ref{pro:1}, our goal is to learn attention masks such that there is minimum variation between the expert policy (\ref{eqn:expert policy}) and mask policy (\ref{eqn:mask policy}). Moreover, Property \ref{pro:2} requires the learned mask to attend to as little information as possible. These considerations lead to the mask loss function as follows:
 \begin{shrinkeq}{-0.2ex}
 \begin{align}
  \label{eqn:mask loss}  L_\text{mask} = \sum_{i=1}^{N} \frac{1}{2}\big\|\pi_m(\bs_i)-\pi_e(\bs_i)\big\|^2_2 + \alpha \left\|f^\theta_\text{exp}(\bs_i)\right\|_1,
 \end{align}
 \end{shrinkeq}
 where $\|\cdot\|_1$ denotes the $L_1$-norm, $\alpha$ is a positive scalar controlling the sparseness of the mask, and $N$ is the batch size. The first term ensures that the agent's behaviour does not change much after overlaying the state with corresponding attention mask, while the second term is a sparse regularization that pushes for better visual explanations for humans.

 One point worth noting is that only the \emph{decoder} $f_d$ is trained in the second stage, since the \emph{encoder} $f_e$ of SSINet is shared from the \emph{feature extractor} of actor network, which is pretrained in the first stage. The pseudo-code of our training procedure is summarized in Algorithm \ref{alg:sma2}.

%------------------------------------------------------------------------
\section{Validity of Our Method}\label{sec:experiment}
 Before we apply the proposed method to provide explanations for vision-based RL, we first verify the effectiveness of our method through performance evaluation and comparative evaluation in this section. Then, our method is applied to render empirical evidences about how the agent makes decisions in Section \ref{sec:applications} and why the agent performs well or badly in Section \ref{sec:saliency metric}. The code and implementation details are available at: https://github.com/shiwj16/SSINet.

%------------------------------------------------------------------------
\subsection{Setup}\label{subsec:env}
 We verify and evaluate the performance of our method on several Atari 2600 games and Duckietown environment (see below for details). All experimental details are given in Appendix \ref{app:details} of the supplemental material. Note that during data collection, the expert policy $\pi_e$ is used to generate $M=50000$ state-action pairs for training the SSINet.

\textbf{Duckietown.}
 Duckietown \cite{gym_duckietown} is a self-driving car simulator environment for OpenAI Gym \cite{openai2016gym}. It places the agent, a Duckiebot, inside of an instance of a Duckietown: a loop of roads with turns, intersections, obstacles, and so on. Specifically, the states are selected as $120 \times 160 \times 3$ color images which are resized from single camera images with $480 \times 640 \times 3$, and the actions contain two continuous normalized numbers corresponding to forward velocity and steering angle respectively. The goal of an agent is to drive forward along the right lane, hence the agent will be rewarded for being as close as possible to the center line of the lane, and also for facing the same direction as the lane's tangent. In our experiments, we evaluate our method on Lane-following task, and the $\emph{empty}$ map is chosen as the training scene. In addition to \emph{empty} and \emph{zigzag} maps provided by the official environment, another eight customized maps (including \emph{empty-city}, \emph{zigzag-city}, \emph{corner}, \emph{corner-city}, \emph{U-turn}, \emph{U-turn-city}, \emph{S-turn} and \emph{S-turn-city}) are designed only for evaluation. These ten maps are mainly different from each other in background and driving route, detailed descriptions are given in Appendix \ref{app:maps} of the supplemental material.

\textbf{Atari 2600.}
 In addition to Duckietown, we also perform experiments in the Arcade Learning Environment (ALE) \cite{bellemare2013arcade}, which is a commonly used benchmark environment for discrete RL tasks. In our experiments, we select several challenging games, such as Tennis, Riverraid and Breakout. In each game, the agent receives $84 \times 84 \times 4$ stacked grayscale images as inputs, as described in \cite{mnih2015human}. It is worth noting that the Riverraid and Breakout tasks have no strong pattern of large ``super pixels''. While the agent uses the boundary for attention as opposed to wide lane on Riverraid task, the task-relevant features on Breakout task have non-trivial patterns, which make super-pixel based methods (such as flood-fill) impossible. Tennis is a singles tennis game where direct implementation of our method may fail, since the task-relevant feature happens to be black. To tackle this problem, we change the color of black features by adding a small value to each pixel of the state when applying our method to the Tennis task.

%------------------------------------------------------------------------
 \begin{figure*}[t]
  \setlength{\abovecaptionskip}{-0.05cm}
  \setlength{\belowcaptionskip}{-0.4cm}
  \begin{center}
   \includegraphics[width=0.88\linewidth]{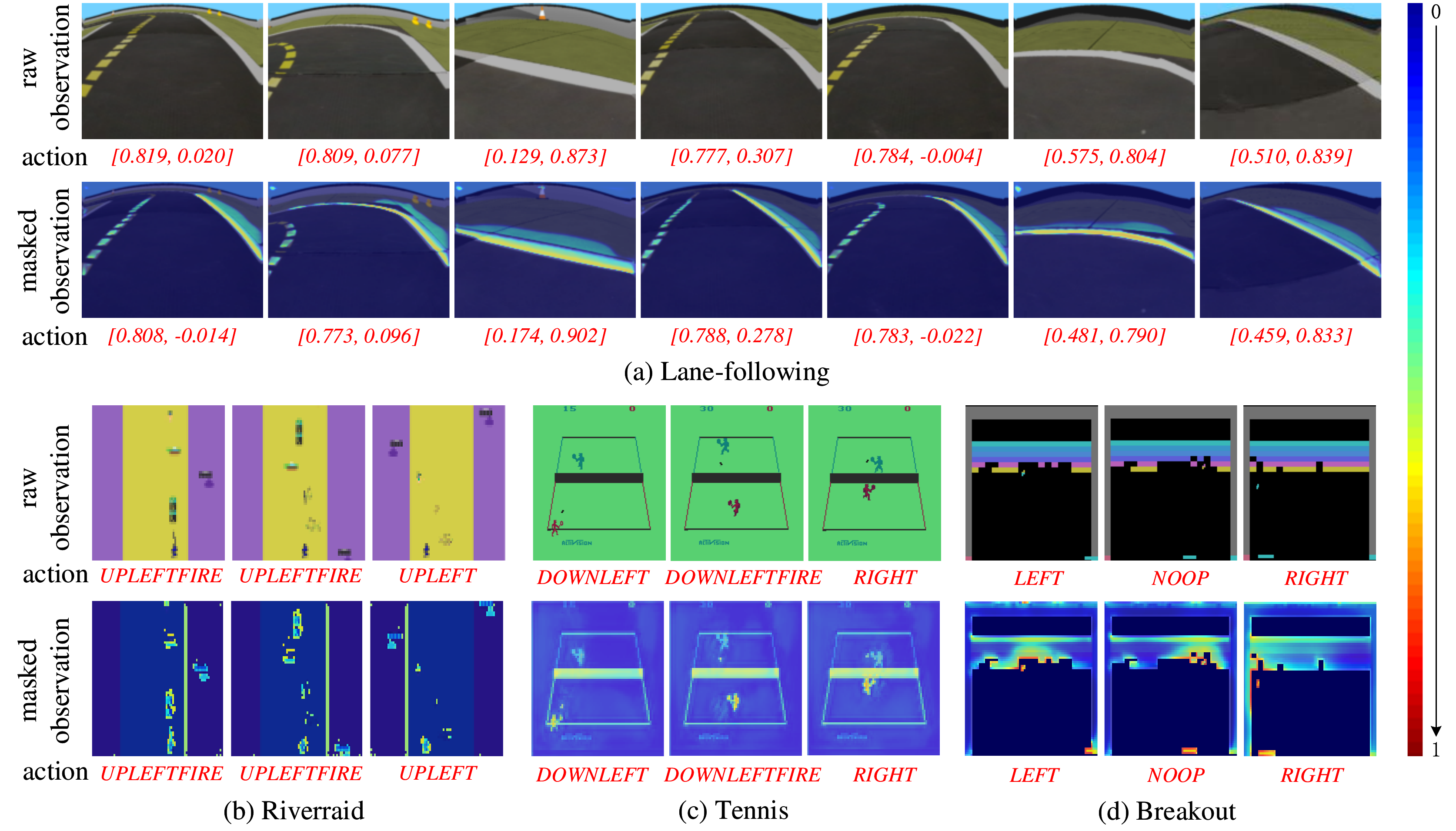}
  \end{center}
  \captionsetup{font={small}}
  \caption{Basic attention patterns (time goes from left to right). For each task, the top row shows a sequence of state-action pairs generated by the expert policy while the bottom row shows a sequence of masked state-action pairs generated by the mask policy. Bright areas in heatmaps are the regions used to make decisions by the mask policy. Best viewed on a computer monitor.}
  \label{fig:basic mask patterns}
 \end{figure*}
\subsection{Evaluations}\label{sec:validity}
 To demonstrate the effectiveness of our proposed method, we verify the RL agent's behaviour consistency between mask policy and expert policy from two aspects, i.e., \emph{average return} and \emph{behavior matching}. \emph{Average return} represents the long-term rewards of two policies, while \emph{behavior matching} characterizes the behavioural similarity of two policies. We note that similar metrics are also used for attention-guided learning in recent work \cite{zhang2019atari}.

 Fig. \ref{fig:basic mask patterns} shows the results of our SSINet on four tasks in terms of \emph{behavior matching}. It can be observed that the mask policy makes decisions using partial information (or the bright areas in the heatmaps) while the expert policy uses all information in the state, but as expected, the mask policy predicts almost the same actions as the expert policy. This observation verifies that the attention masks produced by our SSINet can accurately highlight the task-relevant information, which constitutes most evidence for the expert policy's behaviour.
 \begin{figure}[t]
  % \vspace{-3ex}
  \setlength{\abovecaptionskip}{-0.1cm}
  \setlength{\belowcaptionskip}{-0.2cm}
  \begin{center}
   \includegraphics*[width=0.88\linewidth]{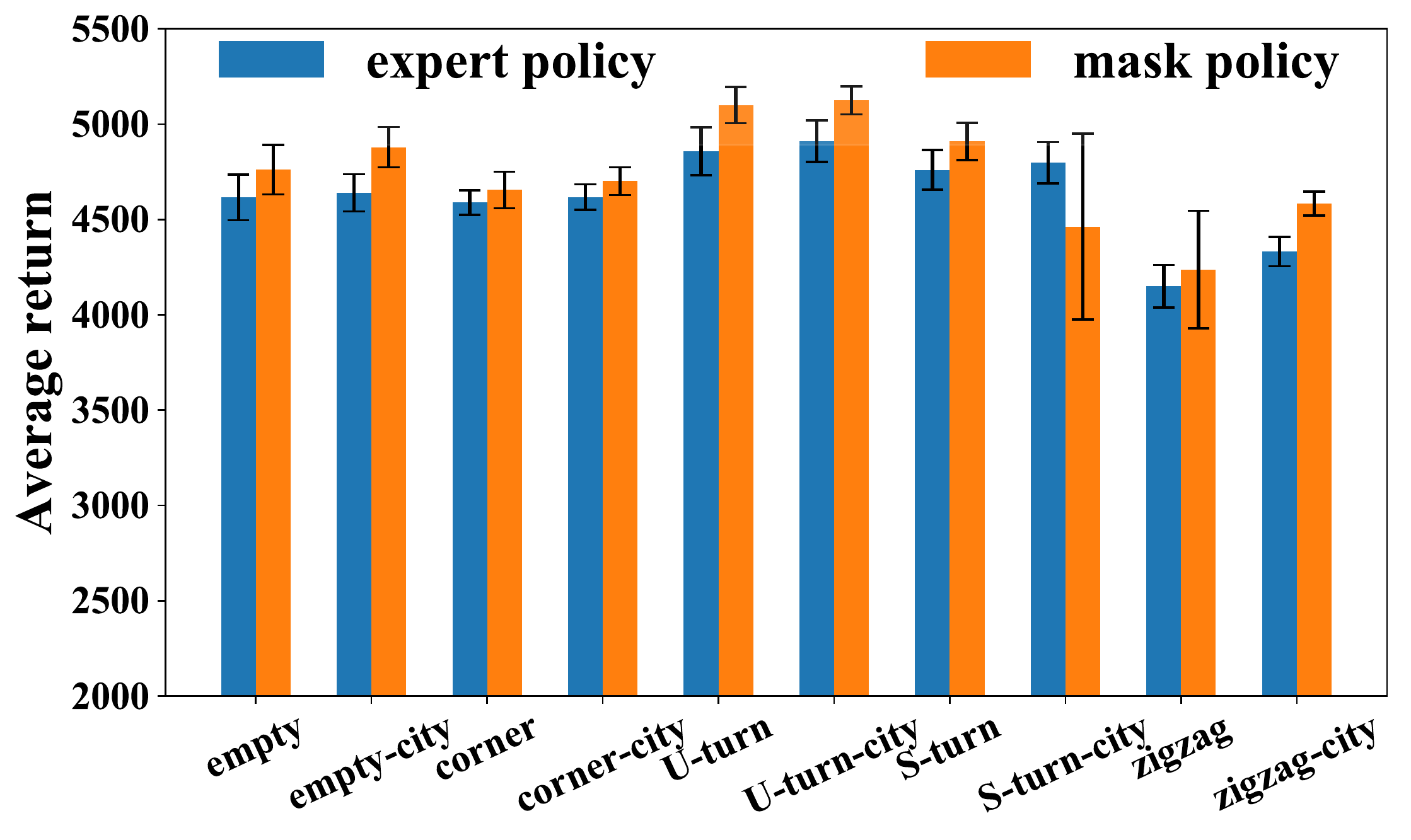}
  \end{center}
  \captionsetup{font={small}}
  \caption{Performance comparisons between expert policy and mask policy on Lane-following task. Both two policies are evaluated on ten maps. All returns are averaged across 15 evaluation episodes, and the errorbars represent the standard variations.}
  \label{fig:return}
 \end{figure}

 \begin{figure}[t]
  % \vspace{-3ex}
  \setlength{\abovecaptionskip}{-0.1cm}
  \setlength{\belowcaptionskip}{-0.2cm}
  \begin{center}
   \includegraphics*[width=0.88\linewidth]{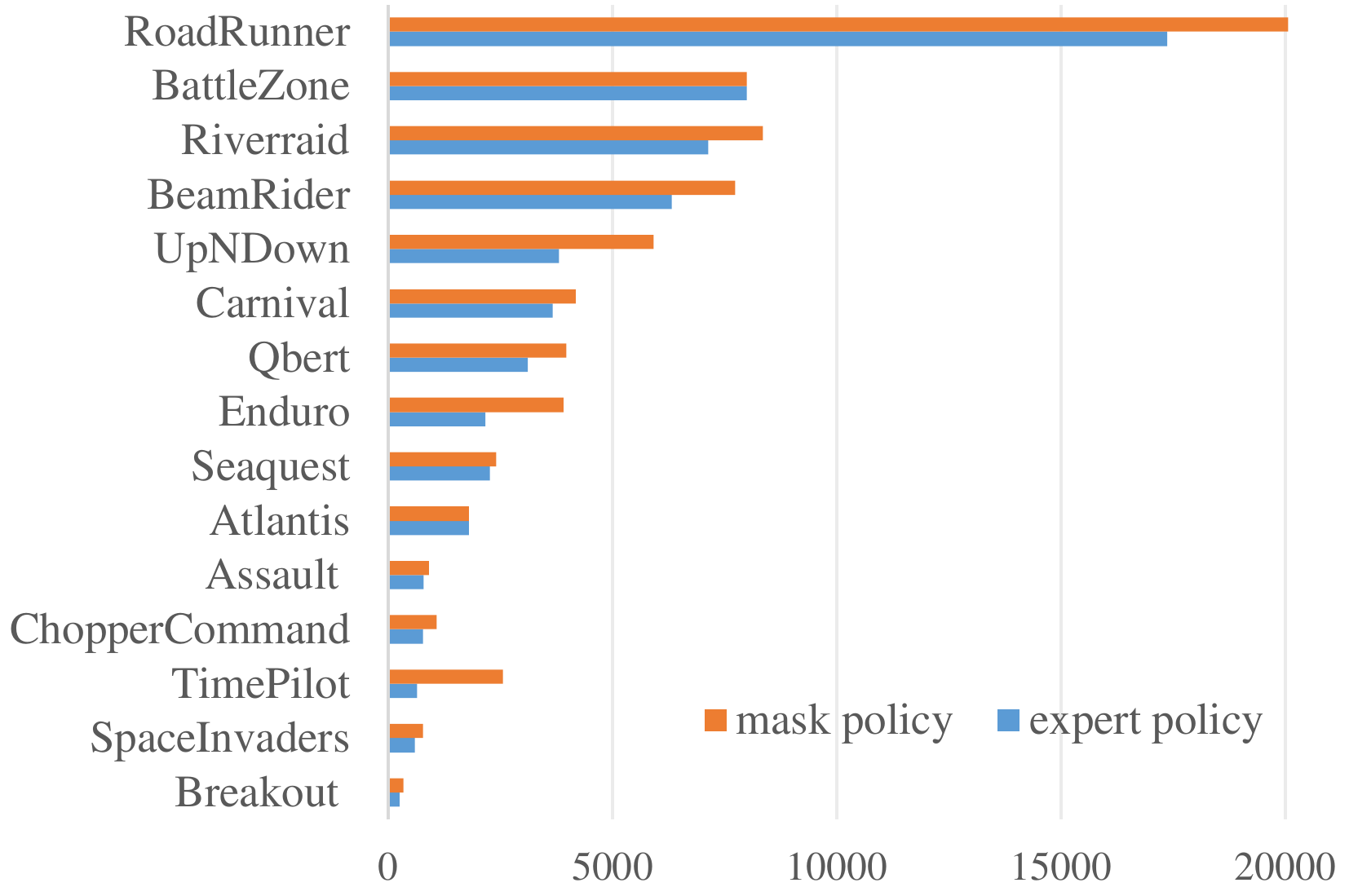}
  \end{center}
  \captionsetup{font={small}}
  \caption{Performance comparisons between expert policy and mask policy on Atari games.}
  \label{fig:return_atari}
 \end{figure}

\begin{figure}[t]
  \setlength{\abovecaptionskip}{-0.1cm}
  \begin{center}
   \includegraphics*[width=0.86\linewidth]{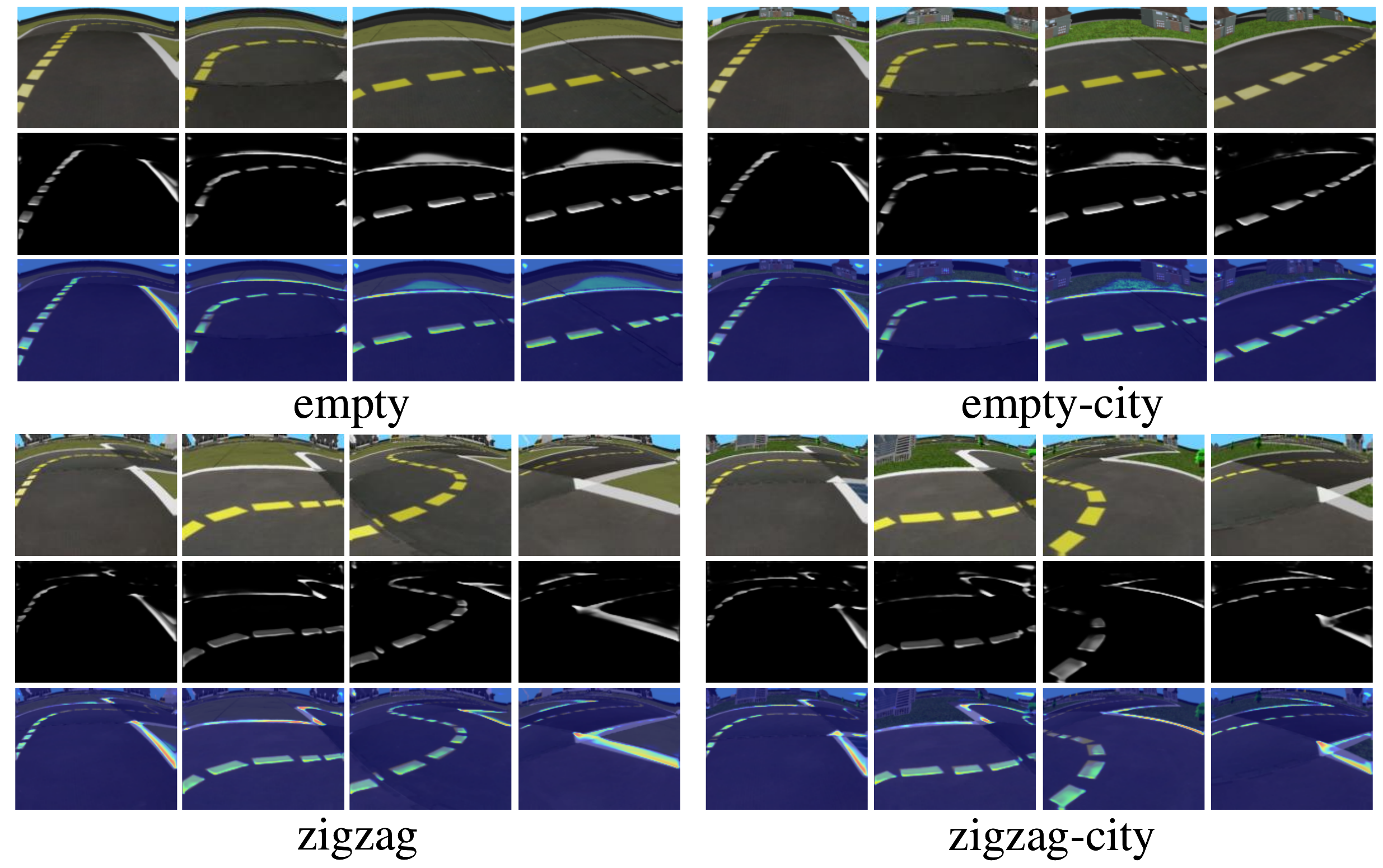}
  \end{center}
  \captionsetup{font={small}}
  \caption{Visualization of the performance on four maps. For each map, three rows represent a sequence of states, attention masks and masked states respectively (time goes from left to right).}
  \label{fig:empty_zigzag}
 \end{figure}

 To further verify the effectiveness of our method, we compare the performance of expert policy and mask policy in terms of \emph{average return}, as shown in Fig. \ref{fig:return} and Fig. \ref{fig:return_atari}. It can be seen that the mask policy consistently achieves greater long-term rewards than the expert policy on all maps except the \emph{S-turn-city} map. As stated in Section \ref{sec:sma2}, the expert policy and mask policy take as input original state and attention-overlaid state, respectively. Therefore, we can conclude that the attention masks produced by SSINet can retain task-relevant information and filter out task-irrelevant information. Fig. \ref{fig:empty_zigzag} visualizes the performance on \emph{empty}, \emph{empty-city}, \emph{zigzag} and \emph{zigzag-city} maps, and more visualization results are given in Appendix \ref{app:additional results} of the supplemental material.

\begin{figure*}[t]
  \setlength{\abovecaptionskip}{-0.1cm}
  \setlength{\belowcaptionskip}{-0.25cm}
  \begin{center}
   \includegraphics*[width=0.86\linewidth]{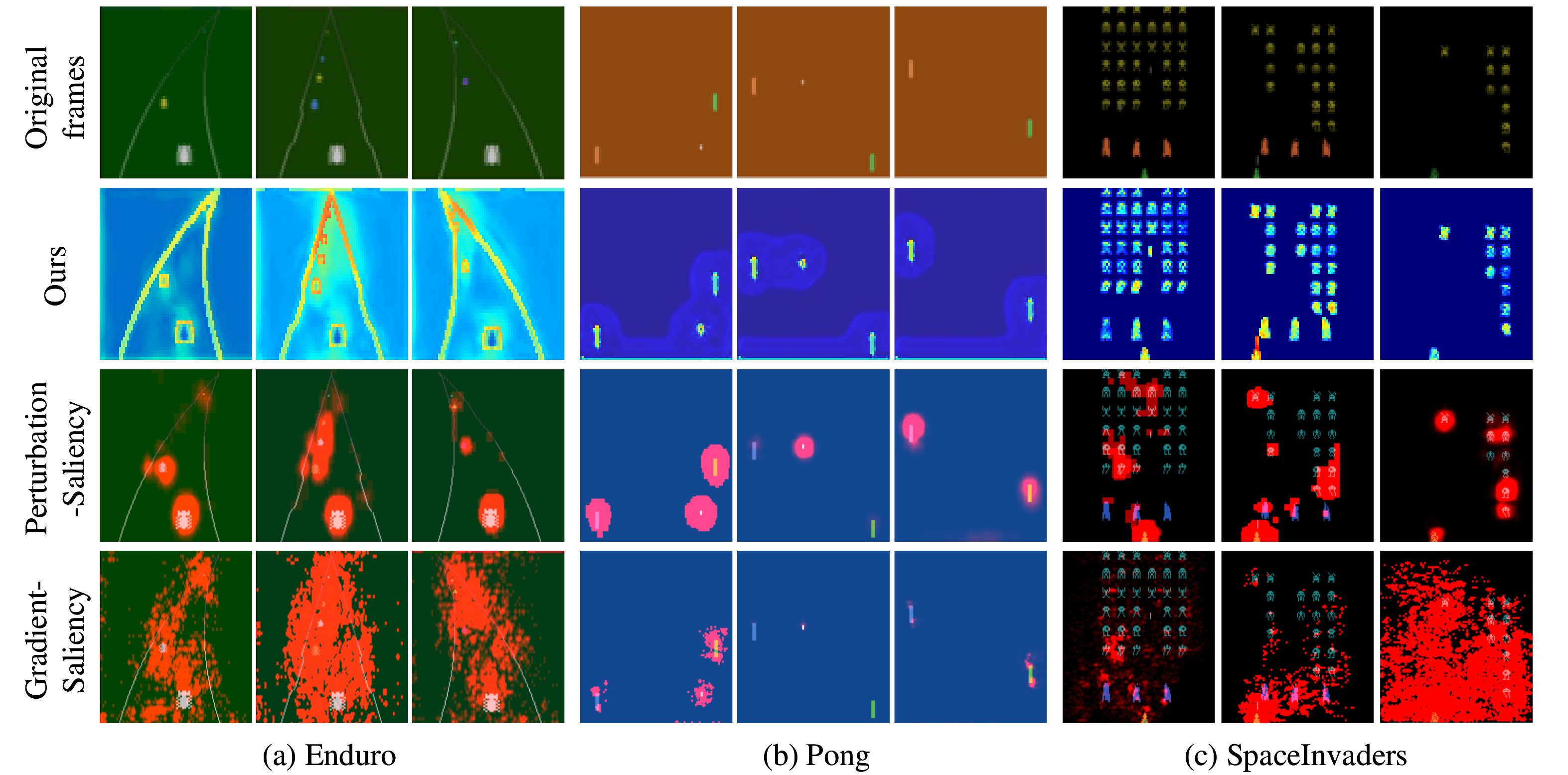}
  \end{center}
  \captionsetup{font={small}}
  \caption{Comparison between our method and other methods, including Gradient-based saliency method (Gradient-Saliency) \cite{zahavy2016graying} and Gaussian perturbation-based saliency method (Perturbation-Saliency) \cite{greydanus2018visualizing}. For each method, a short sequence of saliency-overlaid states is shown.}
  \label{fig:comparisons}
 \end{figure*}

 \textbf{Comparative evaluation.}
 We compare our method against several popular RL explanation methods including Gradient-based saliency method (Gradient-Saliency) \cite{zahavy2016graying} and Gaussian perturbation-based saliency method (Perturbation-Saliency) \cite{greydanus2018visualizing}. These methods focus on the interpretability of vision-based RL and have been briefly reviewed in Section \ref{sec:related work}. Fig. \ref{fig:comparisons} compares our method with Jacobian-based and Perturbation-based methods in terms of the quality of saliency map. The results show that our method produces higher-resolution and sharper saliency maps than the others. Moreover, the saliency maps generated by our method can reflect the relative importance of features by the depth of color. In contrast, the other methods can only locate the position of features. Note that the blue in the last two rows of Fig. \ref{fig:comparisons} does not correspond to saliency value.

 To quantitatively evaluate the quality of masks generated by different methods, we further compare the average return of several policies, which can access to only the pixels of particular masks obtained by different methods during the training. Table \ref{tab:comparison return} summarizes the average return of three different methods, and all results are averaged across five random training runs. It can be seen that the policy still achieves good performance when accessing to only the pixels of the mask generated by our method. In contrast, we can observe obvious performance degradation when using the other methods to generate the masks.

%------------------------------------------------------------------------
\section{Interpreting the Decision of Agents}\label{sec:applications}
 In this section, our method is applied to explain \emph{how RL agent makes decisions}. First, basic attention patterns for making decisions are recognized and understood. Second, the relative importance of different task-relevant features is identified for easy understanding of the agent's decision-making process. Third, some failure cases are analyzed from the viewpoint of \emph{attention shift}.

%------------------------------------------------------------------------
\subsection{Basic Attention Patterns for Making Decisions}
 Here we explain how RL agent makes decisions by understanding the agent's basic attention patterns. As visualized in Fig. \ref{fig:basic mask patterns}, the most dominant pattern is that the agent focuses on only small regions which are strongly task-relevant, while other regions are very ``blurry'' and can be ignored. In other words, the state is not a primitive, the agent learns what information is important for making decisions and where to look at each time step. For example, the task-relevant features are white edge line and yellow dashed line on Lane-following task, players, tennis net and ball on Tennis task. In fact, this conclusion is consistent with human gaze-action pattern \cite{land2009vision}, one characteristic of which is that humans tend to focus attention selectively on parts of the visual space to acquire task-relevant information when and where it is needed.

 \begin{table}[t]
  \centering
  % \fontsize{10.0}{11}\selectfont
  %\setlength{\tabcolsep}{1.5mm}{}
  \caption{The comparison of average returns when the policy is learned with access to only the pixels of particular masks obtained by different methods during the training.}
  \label{tab:comparison return}
  \begin{threeparttable}
   \begin{tabular}{cccc}
    \toprule
     \multirow{2}{*}{Tasks} & \multirow{2}{*}{Ours} & \multirow{2}{*}{\shortstack{Perturbation\\method \cite{greydanus2018visualizing}}} & \multirow{2}{*}{\shortstack{Gradient\\method \cite{zahavy2016graying}}} \cr
                  &                  &         &         \cr
    \midrule
    Enduro        & \textbf{2755.42} & 1741.10 & 819.92  \cr
    Seaquest      & \textbf{2356.67} & 1830.00 & 836.00  \cr
    SpaceInvaders & \textbf{740.45}  & 562.92  & 297.73  \cr
    \bottomrule
   \end{tabular}
  \end{threeparttable}
 \end{table}

%------------------------------------------------------------------------
\subsection{Relative Importance of Task-Relevant Features}\label{sec:importance of soft mask}
 In addition to making it clear what information is used and where to look, understanding the relative importance of different task-relevant features is also crucial for easily explaining the agent's decision-making process. Although the value of attention mask (or the depth of color in heatmap) has intuitively indicated the relative importance of different features in the state, it is not strictly verifiable.

\begin{figure}[t]
  \setlength{\abovecaptionskip}{-0.05cm}
  \setlength{\belowcaptionskip}{-0.2cm}
  \begin{center}
   \includegraphics*[width=0.90\linewidth]{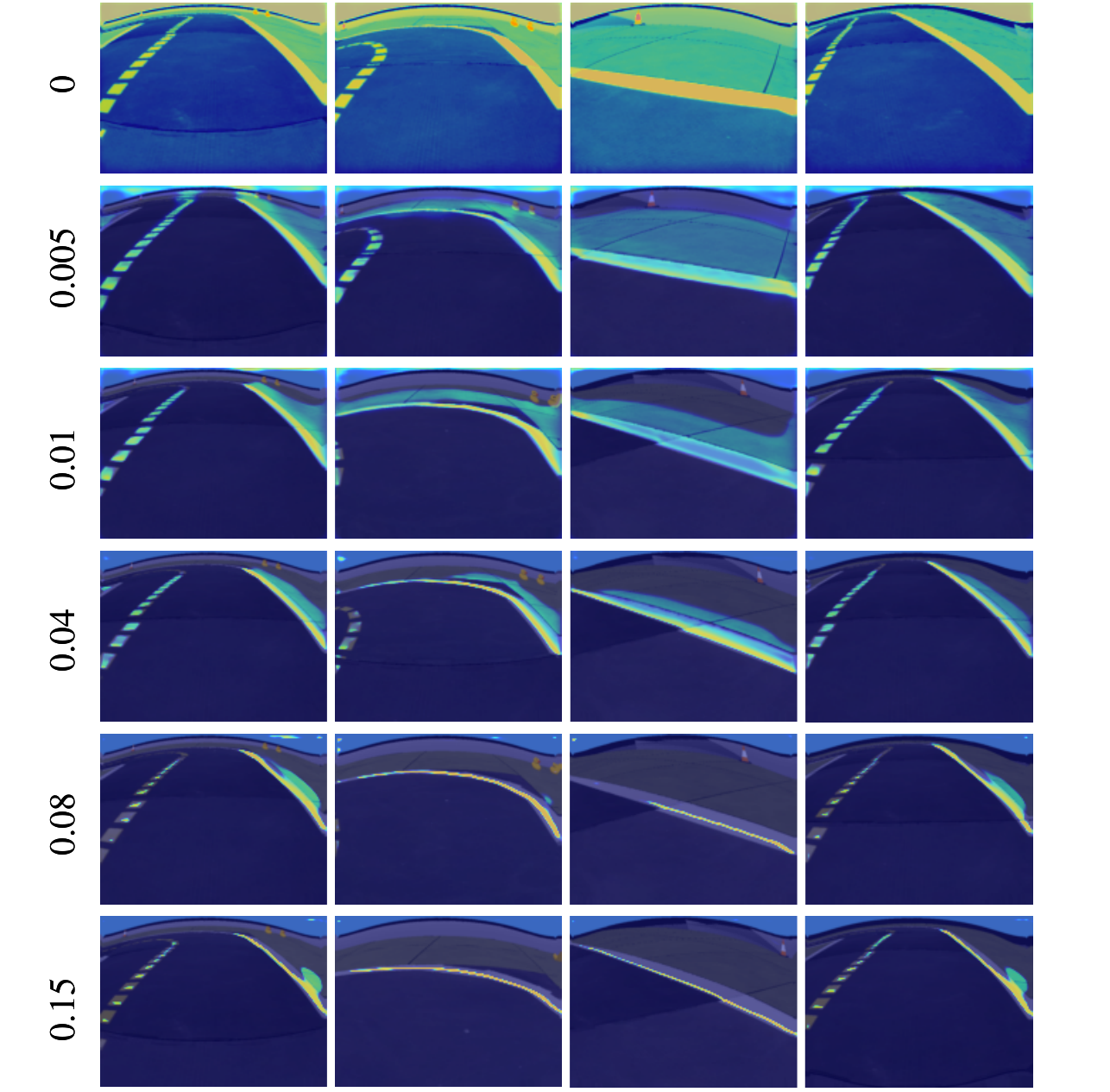}
  \end{center}
  \captionsetup{font={small}}
  \caption{Evolution of the agent's attention as the regularization scale $\alpha$ varies. Bright areas are the ``attended'' regions for making decisions.}
  \label{fig:reg scale}
 \end{figure}

 In this section, we seek to identify the relative importance of task-relevant features in a more interpretable way. We observe that greater regularization scale $\alpha$ in mask loss (\ref{eqn:mask loss}) actually means severer penalty to the agent for attending to task-irrelevant regions. Based on this observation, we propose to assess the relative importance of task-relevant features by comparing multiple attention masks trained with different values of $\alpha$. To that end, we perform a fine search to visualize the evolving process of attention masks. Fig. \ref{fig:reg scale} shows the evolution of attention masks in the form of heatmap as the regularization scale $\alpha$ varies.

 As can be seen in Fig. \ref{fig:reg scale}, with the increasing of regularization scale, the ``attended'' regions are gradually narrowed down to the most important information as expected. Concretely, the inner side of both yellow dashed line and white edge line is considered to be more important than the outer side for making decisions, and nearby lines are considered to be more important than distant lines. In fact, this conclusion is consistent to human gaze system where limited visual sensor resources will be assigned to the most important information.

%------------------------------------------------------------------------
\vspace{-0.5ex}
\subsection{Analysis of Failure Case}
 In practice, it is critical to ensure that a trained RL agent can be directly transferred to novel scenes different from the scene for training. However, robustness is not always guaranteed. Take Lane-following task for example, as can be seen in Fig. \ref{fig:return}, there is a significant performance degradation when transferring the agent trained on \emph{empty} map to other maps, such as \emph{S-turn-city}, \emph{zigzag} and \emph{zigzag-city}. This robustness problem can be explained intuitively from the point of view of \emph{attention shift}.

\begin{figure}[t]
  \setlength{\abovecaptionskip}{-0.05cm}
  \setlength{\belowcaptionskip}{-0.3cm}
  \begin{center}
   \includegraphics*[width=0.9\linewidth,viewport=152 138 705 420]{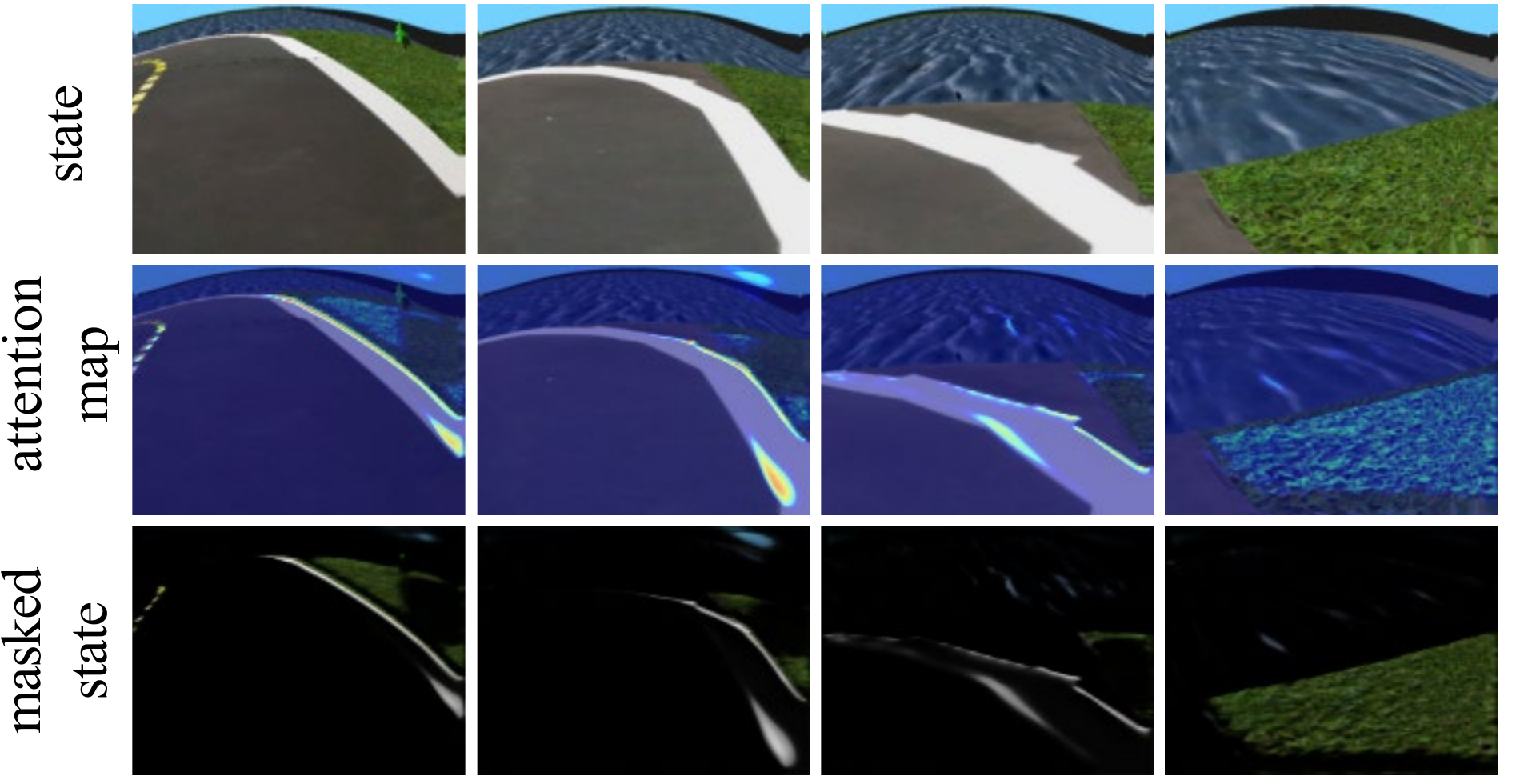}
  \end{center}
  \captionsetup{font={small}}
  \caption{A problematic situation (time goes from left to right). Three rows correspond to original states, generated attention maps and masked states, respectively.}
  \label{fig:robustness}
 \end{figure}

 In those failure cases, we notice that the agent is prone to divert its attention from task-relevant information to background when facing some novel situations, this phenomenon is called \emph{attention shift}. Fig. \ref{fig:robustness} visualizes a common problematic situation leading to poor robustness in \emph{S-turn-city} and \emph{zigzag-city} maps, in which the agent needs to turn left on a corner surrounded by the grassland and lake. However, this novel situation has never been encountered on \emph{empty} map when training, hence it may be difficult for the agent to judge what features are important for making decisions in current situation. As a result, the agent gradually loses attention to task-relevant information (white edge line) and mistakenly attends to the background (lake and grassland), then it gets into stuck due to catastrophic cumulative attention shift.

%------------------------------------------------------------------------
\vspace{-0.5ex}
\section{Interpreting the Performance of Agents}\label{sec:saliency metric}
 In this section, our method is applied to explain \emph{why the agent performs well or badly} quantitatively, especially when transferred to novel scenes. To that end, we start with introducing two evaluation metrics to assess the attention masks generated by SSINet. These metrics allow us to give quantitative explanation about how the agent's attention influences its performance. Then, they are further used to explain the performance of RL agents trained with different algorithms and actor architectures. Finally, potential extension of our method to self-supervised learning is briefly discussed.

%------------------------------------------------------------------------
\subsection{Mask Evaluation Metrics}
 To interpret the performance of RL agents, two evaluation metrics are introduced to assess the quality of generated attention masks. Specifically, \emph{feature overlapping rate} and \emph{background elimination rate} are defined as:
 \begin{itemize}
  \item \emph{Feature Overlapping Rate} (FOR) - the overlapping ratio between the area of true mask and learned mask.
  \item \emph{Background Elimination Rate} (BER) - the ratio of eliminated background area by the mask to the whole background area.
 \end{itemize}

 \begin{figure}[t]
  \setlength{\abovecaptionskip}{-0.03cm}
  \setlength{\belowcaptionskip}{-0.3cm}
  \centering
  \subfigure[state]{
   \begin{minipage}[t]{0.28\linewidth}
    \centering
    \includegraphics[width=0.90in]{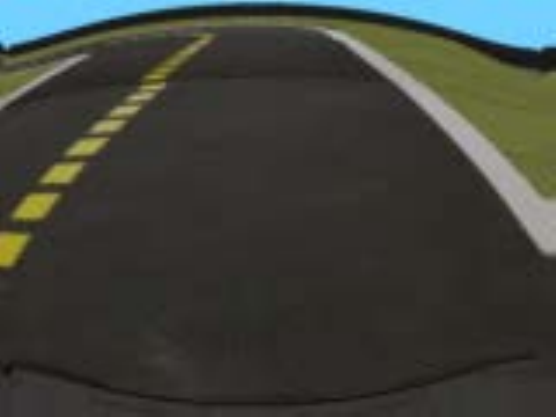}
   \end{minipage}
   \label{subfig:example obs}
  }
  \subfigure[``true'' feature]{
   \begin{minipage}[t]{0.28\linewidth}
    \centering
    \includegraphics[width=0.95in]{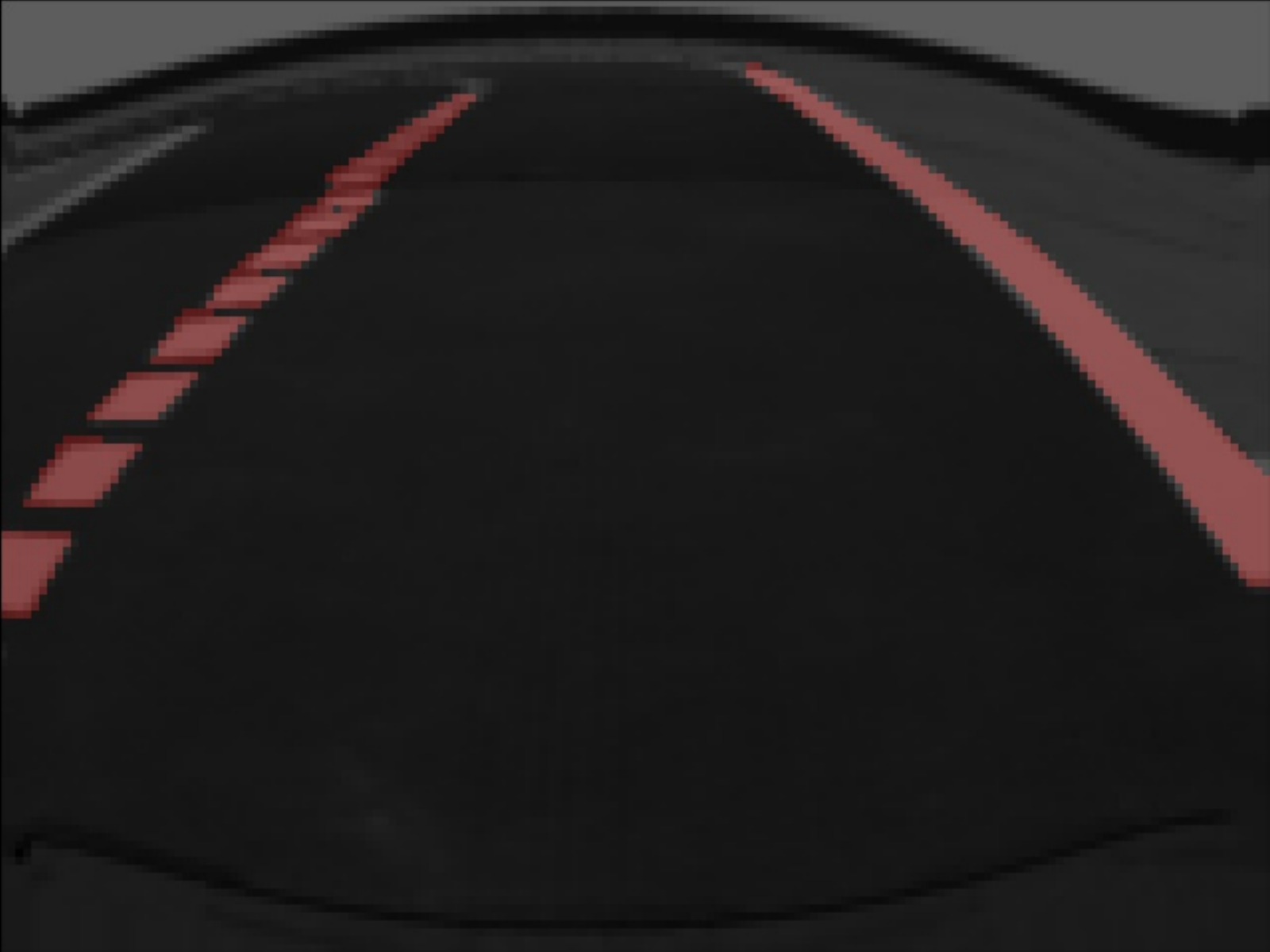}
   \end{minipage}
   \label{subfig:example true feature}
  }
  \subfigure[extracted feature]{
   \begin{minipage}[t]{0.28\linewidth}
    \centering
    \includegraphics[width=0.95in]{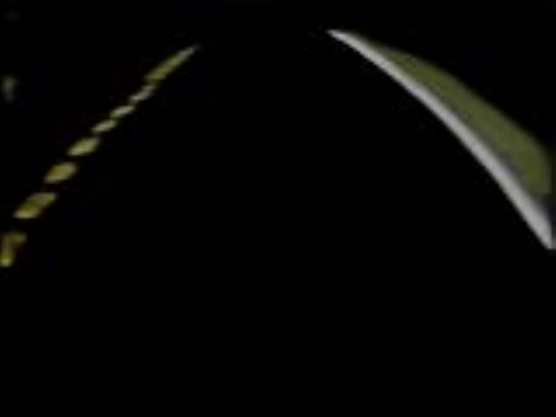}
   \end{minipage}
   \label{subfig:example extracted feature}
  }
  \centering
  \captionsetup{font={small}}
  \caption{An example for calculating the mask metrics. (a) is a specific state from Duckietown, red area in (b) is the ``true'' task-relevant features that we expect to learn, and unmasked area in (c) is the features extracted by our SSINet.}
  \label{fig:example feature}
 \end{figure}

 For a specific state $\bs$, mask metrics FOR($\bs$) and BER($\bs$) are calculated as follows:
 \begin{shrinkeq}{-0.75ex}
  \begin{align}
   \label{eqn:FOR} \text{FOR}(\bs)&= \frac{S_{e,f}\bigcap S_{t,f}}{S_{t,f}}, \\
   \label{eqn:BMR} \text{BER}(\bs)&= \frac{S_{t,b}-S_{t,b}\bigcap S_{e,f}}{S_{t,b}},
  \end{align}
 \end{shrinkeq}
 where $\cap$ and $\cup$ are union and intersection operators respectively. $S_{e,f}$, $S_{t,f}$ and $S_{t,b}$ represent the area of extracted features, ``true'' task-relevant features and ``true'' background respectively, as shown in Fig. \ref{fig:example feature}. Note that the ``true'' background $S_{t,b}$ is the area outside the ``true'' features $S_{t,f}$ in Fig. \ref{subfig:example true feature}. In general, FOR indicates how agents can extract useful information from the state and BER indicates how the SSINet can eliminate task-irrelevant information in the state. We point out that our FOR and BER metrics are equivalent to \emph{true positive rate} (TPR) and \emph{true negative rate} (TNR) from the perspective of pixel classification, and can be readily converted to TPR and TNR. Nevertheless, we prefer to use FOR and BER to evaluate the quality of the generated masks in the field of RL, since they have a more intuitive meaning and enable easy understanding of RL agents even for non-experts.

 However, it is difficult to obtain the ground-truth feature, which is required to compute FOR and BER. To address this problem, we select Duckietown as the the main environment for quantitative evaluation, and use prior information to manually annotate ``true'' features that are similar to what humans expect. Specifically, we annotate the right edge line and center dashed line as the ``true'' features due to two reasons. First, the goal of Duckietown tasks is to encourage the agent to drive forward along the right lane, and the agent is rewarded for being as close as possible to the center line of the right lane according to the official guidance \footnote{https://github.com/duckietown/gym-duckietown}, hence it is reasonable to treat the right edge line and center dashed line as the ``true'' features. Second, these ``true'' features are consistent with human driving experience. In our experiments, we use averaged mask metrics $\overline{\text{FOR}}$ and $\overline{\text{BER}}$ on the training map to characterize the attention masks generated with our SSINet.

%------------------------------------------------------------------------
\subsection{How the Agent's Attention Influences Performance}
 In order to analyze quantitatively how vision-based RL agent's attention influences its performance, especially when transferred to novel scenes, we compare the average returns $\overline{R}$ of multiple mask policies. These mask policies are all trained to interpret identical RL agent but produce different attention masks for the same state. In our experiments, we train SSINet under different regularization scales $\alpha$ for the same actor network.
 \begin{figure}[t]
  \begin{center}
   \includegraphics*[width=0.9\linewidth]{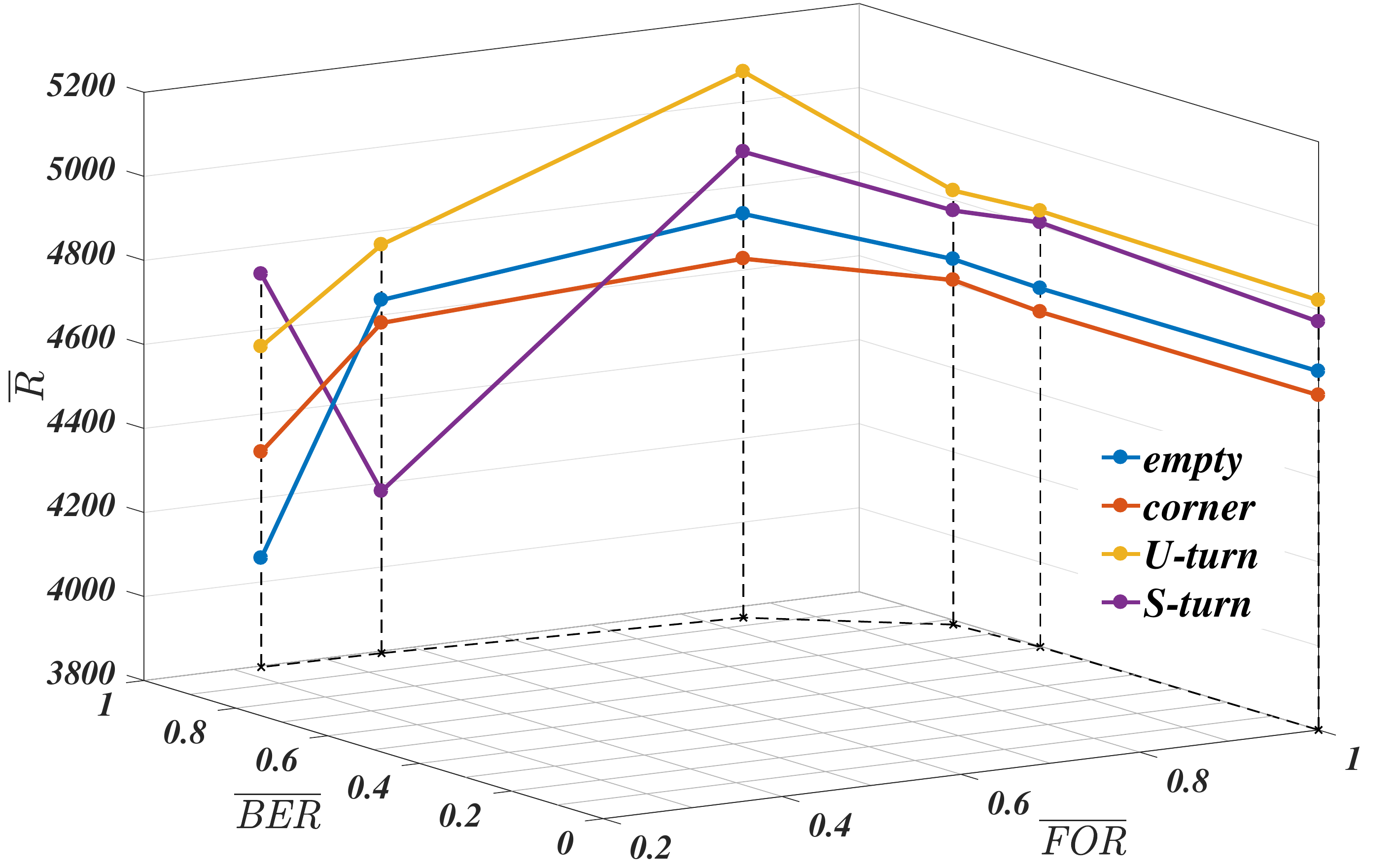}
  \end{center}
  \captionsetup{font={small}}
  \caption{Relationship between the average return $\overline{R}$ and the average mask metrics ($\overline{\text{FOR}}$ and $\overline{\text{BER}}$). $\overline{\text{FOR}}$ and $\overline{\text{BER}}$ are the average results across 10 random states. Each black node in the $\overline{\text{FOR}}-\overline{\text{BER}}$ plane corresponds to a mask policy.}
  \label{fig:for ber return}
 \end{figure}
 \begin{table}[t]
  \centering
  % \fontsize{10.0}{11}\selectfont
  %\setlength{\tabcolsep}{1.5mm}{}
  \caption{$\text{R}^2$ correlation value (Pearson) between the average mask metrics and the average return $\overline{R}$.}
  \label{tab:pearson correlation}
  \begin{threeparttable}
   \begin{tabular}{ccccc}
    \toprule
    \multirow{2}{*}{\shortstack{Correlation value $\text{R}^2$\\(Pearson)}} & \multicolumn{4}{c}{Duckietown tasks} \cr
    \cmidrule(lr){2-5}
    & \emph{empty} & \emph{corner} & \emph{U-turn} & \emph{zigzag} \cr
    \midrule
    $\text{R}^2(\overline{\text{BER}}, \overline{R})$ & 0.086 & 0.081 & 0.006 & 0.017  \cr
    $\text{R}^2(\overline{\text{FOR}}, \overline{R})$ & 0.525 & 0.550 & 0.367 & 0.271  \cr
    $\text{R}^2(\overline{\text{F1}},  \overline{R})$ & 0.021 & 0.018 & 0.148 & 0.161  \cr
    $\text{R}^2(\overline{\text{FOR}}\times\overline{\text{BER}}, \overline{R})$ & 0.655 & 0.699 & 0.760 & 0.504 \cr
    $\text{R}^2(\overline{\text{AUC}}, \overline{R})$ & \textbf{0.674} & \textbf{0.715} & \textbf{0.812} & \textbf{0.559}  \cr
    \bottomrule
   \end{tabular}
  \end{threeparttable}
 \end{table}

 Fig. \ref{fig:for ber return} shows how the average mask metrics ($\overline{\text{FOR}}$ and $\overline{\text{BER}}$) influence the average return $\overline{R}$ on four maps. As can be seen in Fig. \ref{fig:for ber return}, when evaluated on the same map, the agent performs differently with regards to different $\overline{\text{FOR}}-\overline{\text{BER}}$. Only when both $\overline{\text{FOR}}$ and $\overline{\text{BER}}$ have high values, the best performance can be achieved. In other words, the agent can not perform well enough if it neglects task-relevant information or attends too much to the background, reflected either by a small $\overline{\text{FOR}}$ or a small $\overline{\text{BER}}$.

 Table \ref{tab:pearson correlation} presents the $\text{R}^2$ correlation value (Pearson) between the average mask metrics ($\overline{\text{FOR}}$, $\overline{\text{BER}}$, $\overline{\text{F1}}$-score and $\overline{\text{AUC}}$) and the average return $\overline{R}$ on four different Duckietown maps. It can be seen that while there is a weak correlation between single metric ($\overline{\text{FOR}}$ or $\overline{\text{BER}}$) and return, the average return is strongly correlated with $\overline{\text{FOR}}\times\overline{\text{BER}}$ or $\overline{\text{AUC}}$. This observation reveals that $\overline{\text{FOR}}$ and $\overline{\text{BER}}$ are not independent of each other, and an agent can get good performance only when both $\overline{\text{FOR}}$ and $\overline{\text{BER}}$ are great enough. In fact, this conclusion is consistent with Fig. \ref{fig:for ber return}.

%------------------------------------------------------------------------
\subsection{Explaining the Performance of Different Agents}\label{sec:algorithm and network}
 Generally, RL agents may exhibit different performance even on simple tasks. To explain why the agent performs well or badly, especially in terms of stability and robustness when transferred to novel scenes, the above mask metrics are utilized to analyze the behaviour of multiple RL algorithms and actor architectures. Such an analysis can provide explainable basis for the selection of models and actor architectures.

 \begin{figure}[t]
  \vspace{0.5ex}
  \setlength{\abovecaptionskip}{-0.05cm}
  \setlength{\belowcaptionskip}{-0.2cm}
  \begin{center}
   \includegraphics*[width=0.9\linewidth]{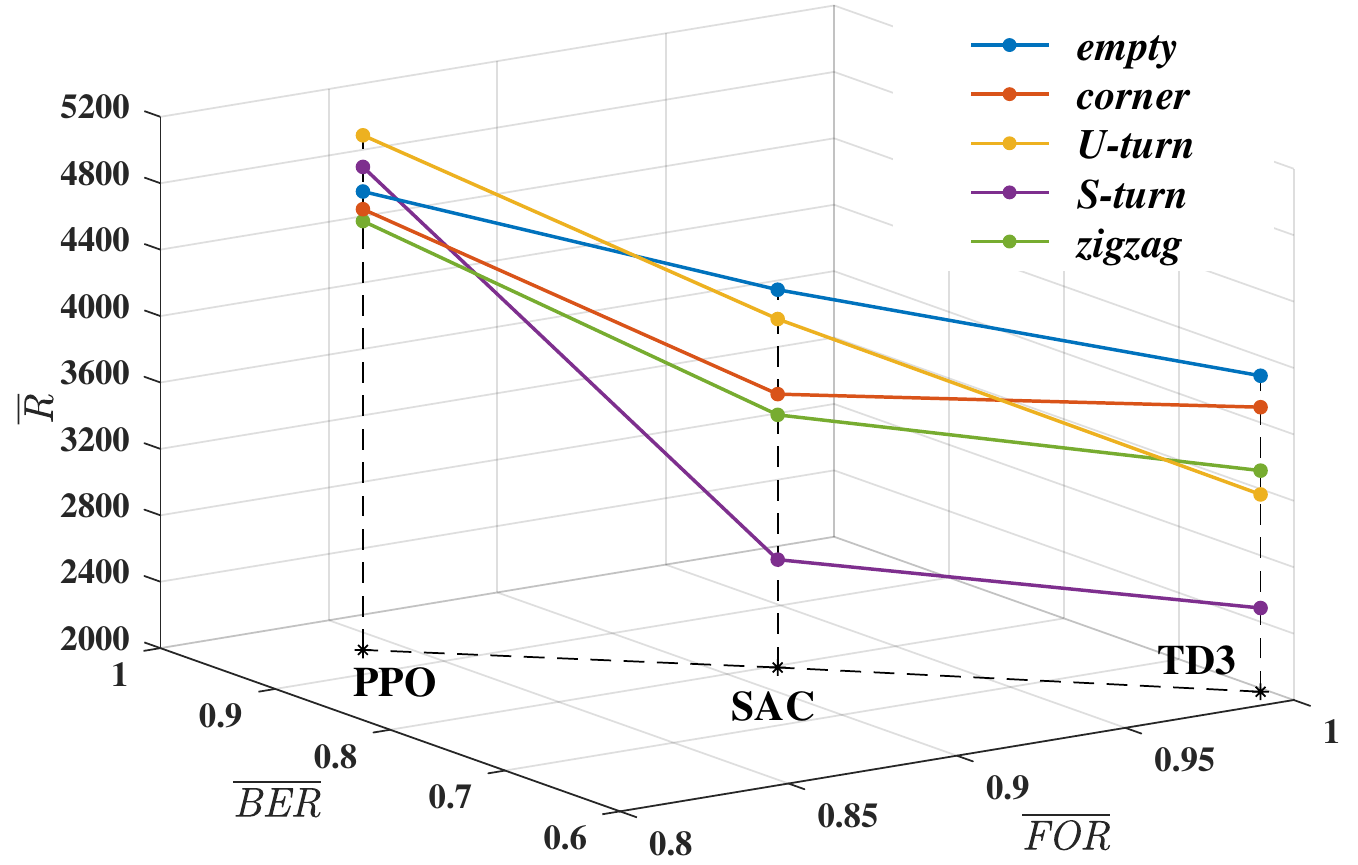}
  \end{center}
  \captionsetup{font={small}}
  \caption{The performance of PPO, SAC and TD3 agents. For each algorithm, the black node in the horizontal plane represents the mask metrics $\overline{\text{FOR}}$ and $\overline{\text{BER}}$, while the vertical axis represents the average return $\overline{R}$ of corresponding mask policy.}
  \label{fig:return of algorithm}
 \end{figure}
 \begin{figure}[t]
  \vspace{1.2ex}
  \setlength{\abovecaptionskip}{-0.1cm}
  \setlength{\belowcaptionskip}{-0.2cm}
  \begin{center}
   \includegraphics*[width=0.9\linewidth]{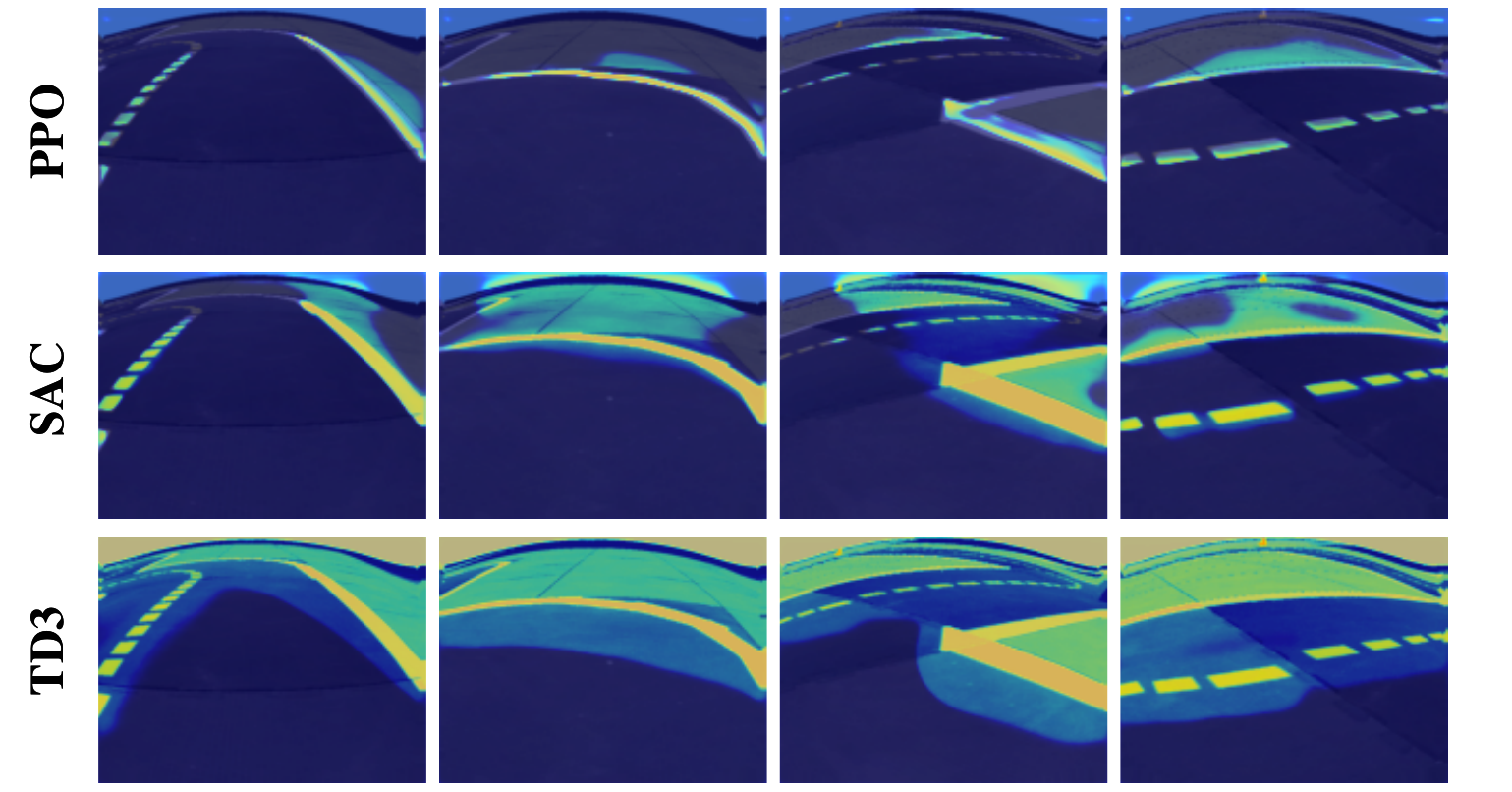}
  \end{center}
  \captionsetup{font={small}}
  \caption{Visualization of performance for PPO, SAC and TD3 algorithms. Scenes in each column are the same.}
  \label{fig:algorithms visualize}
 \end{figure}
\textbf{Case 1: RL algorithm.}
 To explain the performance difference of RL algorithms, we analyze the average return $\overline{R}$ of three popular RL algorithms (PPO, SAC and TD3) with the above mask metrics. As shown in Fig. \ref{fig:return of algorithm}, PPO consistently outperforms both SAC and TD3 on all maps for Lane-following task. The reason for this is the background information has adverse effect on the agent's performance, and our mask metrics can help quantify it. Specifically, although the $\overline{\text{FOR}}$ of SAC and TD3 are close to one indicating that almost all task-relevant information are identified, a large amount of background information is also mistakenly attended to due to small $\overline{\text{BER}}$. In contrast, PPO focuses on main task-relevant information while masking most background information. These conclusions are illustrated and further verified by Fig. \ref{fig:algorithms visualize}, which visualizes the performance of PPO, SAC and TD3. Moreover, we observe that PPO shows better stability than SAC and TD3. Concretely, while PPO agent tends to drive smoothly in the center line of right lane, both SAC and TD3 agents have obvious lateral deviation and drive unsteadily.

\begin{figure}[t]
  \setlength{\abovecaptionskip}{-0.1cm}
  \setlength{\belowcaptionskip}{-0.2cm}
  \begin{center}
   \includegraphics*[width=0.9\linewidth]{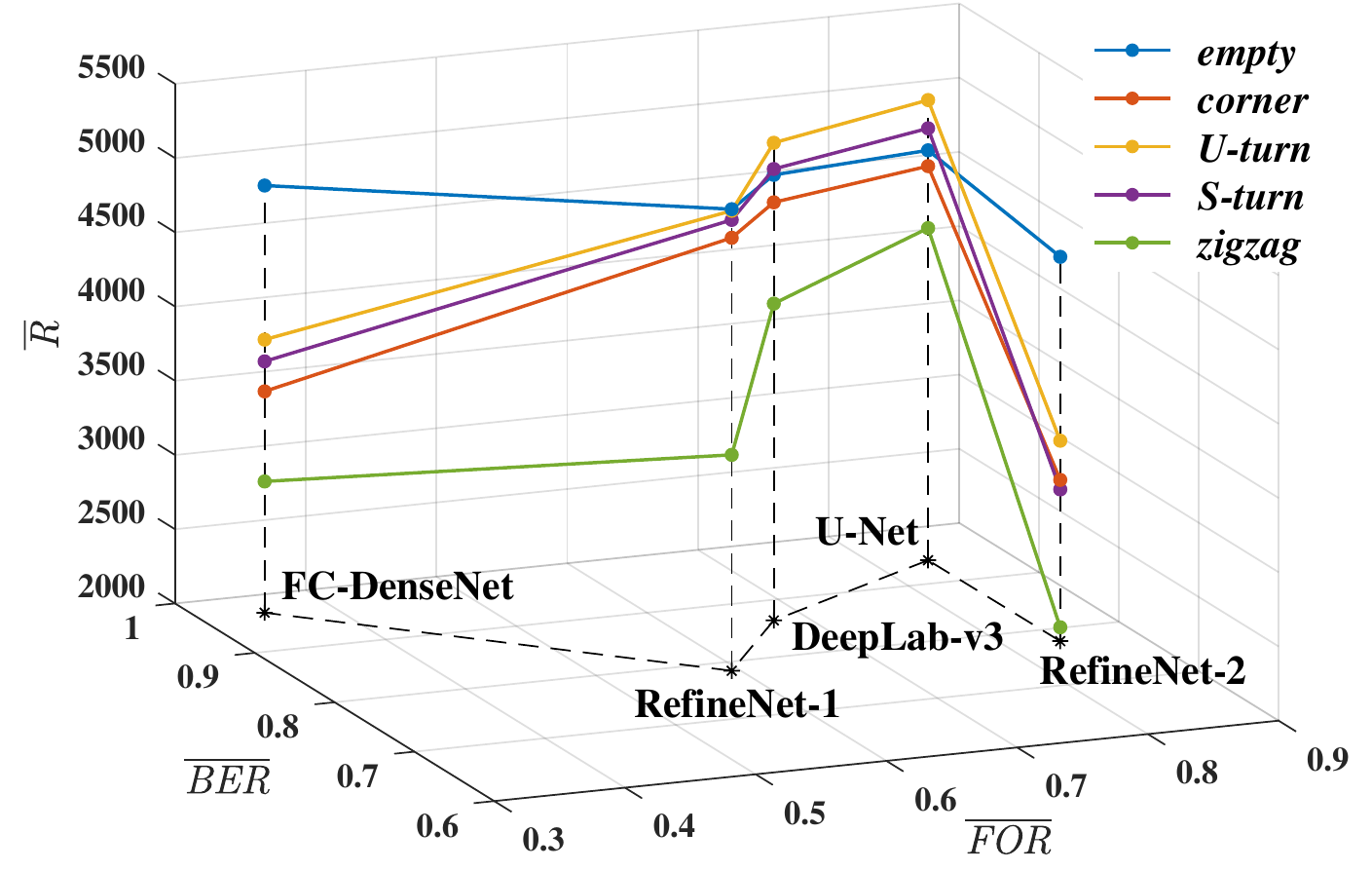}
  \end{center}
  \captionsetup{font={small}}
  \caption{The performance of five agents with different actor network architectures. For each architecture, the black node in the horizontal plane represents the mask metrics $\overline{\text{FOR}}$ and $\overline{\text{BER}}$, while the vertical axis represents the average return $\overline{R}$ of corresponding mask policy.}
  \label{fig:return of network}
 \end{figure}

\begin{figure}[t]
  \vspace{1.2ex}
  \setlength{\abovecaptionskip}{-0.1cm}
  \setlength{\belowcaptionskip}{-0.2cm}
  \begin{center}
   \includegraphics*[width=0.9\linewidth]{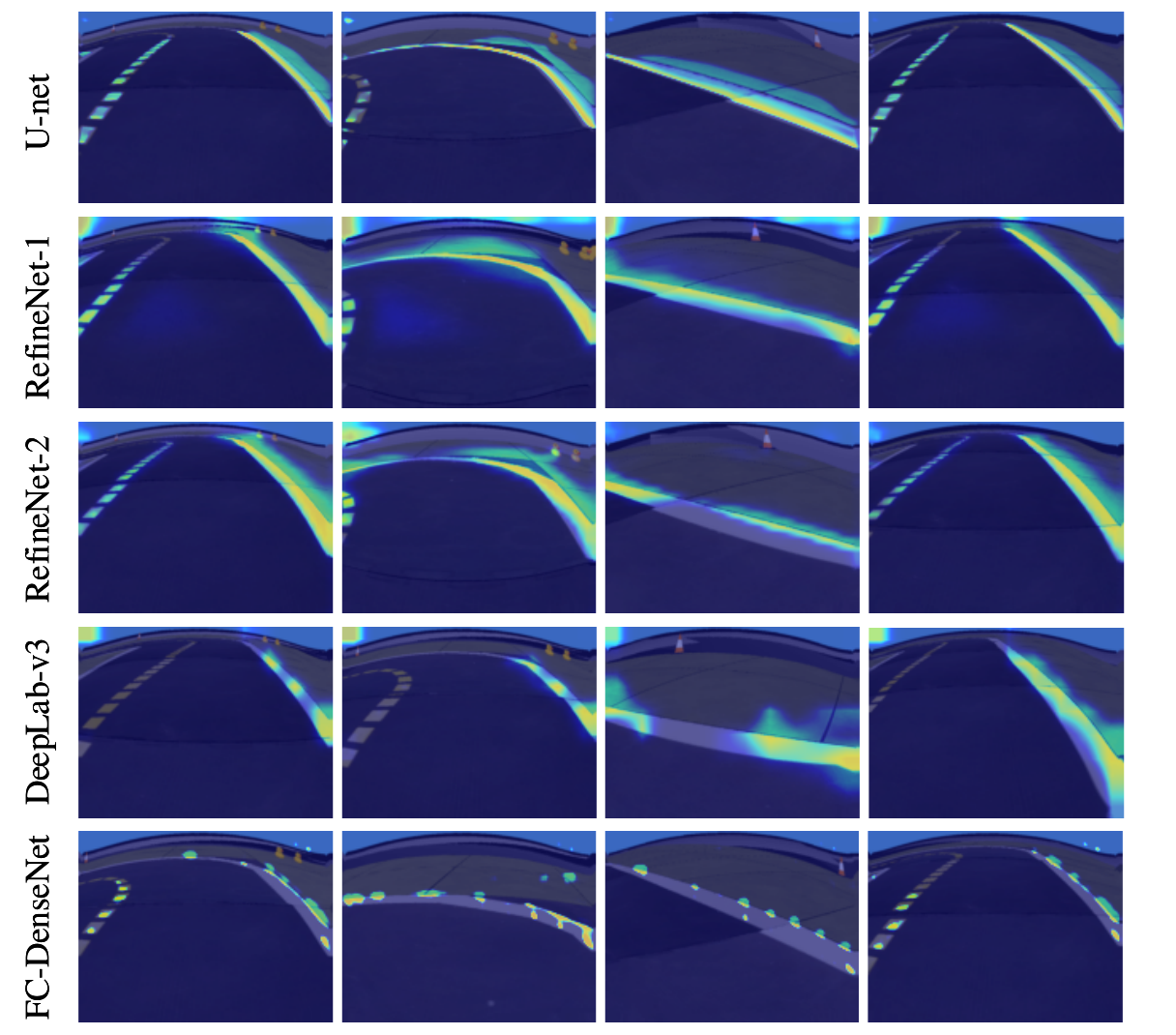}
  \end{center}
  \captionsetup{font={small}}
  \caption{Visualization of performance for Unet, RefineNet, DeepLab-v3 and FC DenseNet. Each row is a sequence of masked states (time goes from left to right). Scenes in each column are similar.}
  \label{fig:networks visualize}
 \end{figure}

\textbf{Case 2: Actor architecture.}
 To understand how the actor architecture affects the agent's performance, we analyze the average return $\overline{R}$ of four popular semantic segmentation architectures (U-Net \cite{ronneberger2015u}, RefineNet \footnote{Notice that RefineNet-1 and RefineNet-2 are exactly the same except for utilizing ResNet and MobileNet as the backbone network respectively.} \cite{lin2017refinenet}, FC-DenseNet \cite{jegou2017one} and DeepLab-v3 \cite{chen2017rethinking}) with the above mask metrics. As shown in Fig. \ref{fig:return of network} and Fig. \ref{fig:networks visualize}, the performance difference among five agents results from diverse attention behaviours indicated by $\overline{\text{FOR}}$ and $\overline{\text{BER}}$. Losing a lot of task-relevant information (corresponding to small $\overline{\text{FOR}}$ such as FC-DenseNet) or mistakenly matching much background (corresponding to small $\overline{\text{BER}}$ such as RefineNet-1 and RefineNet-2) leads to bad performance. Moreover, we conclude that complex actor architecture does not necessarily lead to good performance. In fact, task-relevant features are relatively simple in the context of RL, hence most DNN architectures can extract them well. In practice, small DNNs are generally capable of learning representations and preferred to make the training algorithm focus on the credit assignment problem.

%------------------------------------------------------------------------
\subsection{Potential Extension to Self-Supervised Learning}\label{sec:discussion}
 As a relatively recent learning technique in machine learning, self-supervised learning is challenging due to the lack of labelled data. In our work, we presented a self-supervised interpretable framework for vision-based RL, and a two-stage training procedure is applied to train the SSINet in a self-supervised manner. The learning signal is acquired through the direct interaction between RL agent and environment, and the whole training process is completely label-free. Empirical results in Fig. \ref{fig:basic mask patterns} and Fig. \ref{fig:empty_zigzag} demonstrate that our method is capable of learning high-quality mask through the direct interaction with the environment and without any external supervised signal. In summary, our work may shed light on new paradigms for label-free vision learning such as self-supervised segmentation and detection.

%------------------------------------------------------------------------
\section{Conclusion}
 In this paper, we addressed the growing demand for human-interpretable vision-based RL from a fresh perspective. To that end, we proposed a general self-supervised interpretable framework, which can discover interpretable features for easily understanding the agent's decision-making process. Concretely, a self-supervised interpretable network (SSINet) was employed to produce high-resolution and sharp attention masks for highlighting task-relevant information, which constitutes most evidence for the agent's decisions. Then, our method was applied to render empirical evidences about how the agent makes decisions and why the agent performs well or badly, especially when transferred to novel scenes. Overall, our work takes a significant step towards interpretable vision-based RL. Moreover, our method exhibits several appealing benefits. First, our interpretable framework is applicable to any RL model taking as input visual images. Second, our method does not use any external labelled data. Finally, we emphasize that our method demonstrates the possibility to learn high-quality mask through a self-supervised manner, which provides an exciting avenue for applying RL to self automatically labelling and label-free vision learning such as self-supervised segmentation and detection.

% or

%\appendices

%\section{}

\appendices
%------------------------------------------------------------------------
\section{Network Architectures}\label{app:network architecture}
\begin{figure}[t]
  \setlength{\abovecaptionskip}{-0.05cm}
  \setlength{\belowcaptionskip}{-0.4cm}
  \begin{center}
   \includegraphics*[width=0.95\linewidth]{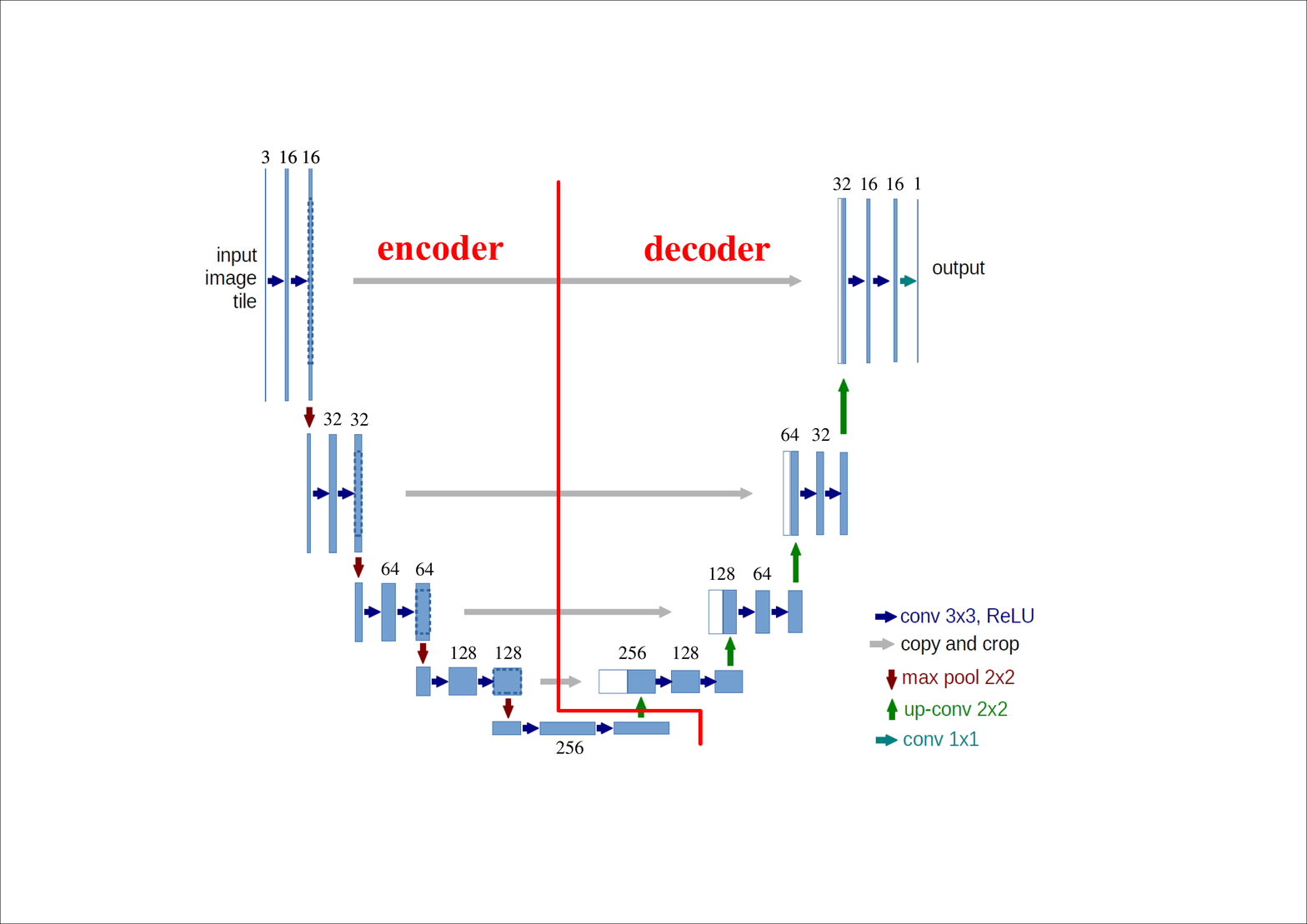}
  \end{center}
  \captionsetup{font={small}}
  \caption{The architecture of our adapted U-Net.}
  \label{fig:unet}
 \end{figure}

 As analyzed in Section 4.2, our SSINet has an encoder-decoder structure. The \emph{encoder} and \emph{decoder} are two convolutional neural networks (CNNs) and together form a general-purpose semantic segmentation network to learn saliency masks. In this paper, four popular semantic segmentation architectures (U-Net, RefineNet, DeepLab-v3 and FC-denseNet) are directly adapted to our SSINet with only minor changes. Detailed design of networks is not our contribution and can be found in corresponding published papers of these semantic segmentation architectures. In our implementation, we only care about which regions are task-relevant and do not need to distinguish those task-relevant features exactly, hence the channel number of final convolution layer is 1, and a sigmoid layer is finally used to output the saliency mask which indicates only task-relevant regions for making decisions. Figure \ref{fig:unet} shows the architecture of our adapted U-Net.

 In our experiments, we use three different algorithms, including PPO, SAC and TD3. Broadly speaking, all these algorithms employ an actor-critic structure. In PPO, the critic is used to predict state value function and thus shares a common feature extractor with the actor. In contrast, both SAC and TD3 employ a separate critic to predict state-action value function. In our implementation, the actor network includes a feature extractor and an action predictor. While the action predictor is a simple two-layer perception, the feature extractor has the same architecture as the encoder of SSINet. For SAC and TD3, the critic network is a separate multi-layer CNN, which is followed by a two-layer perception to predict the state-action value function.

%------------------------------------------------------------------------
\section{Customized Maps on Duckietown}\label{app:maps}
\begin{figure}[t]
  \setlength{\abovecaptionskip}{-0.05cm}
  \begin{center}
   \includegraphics*[width=0.95\linewidth]{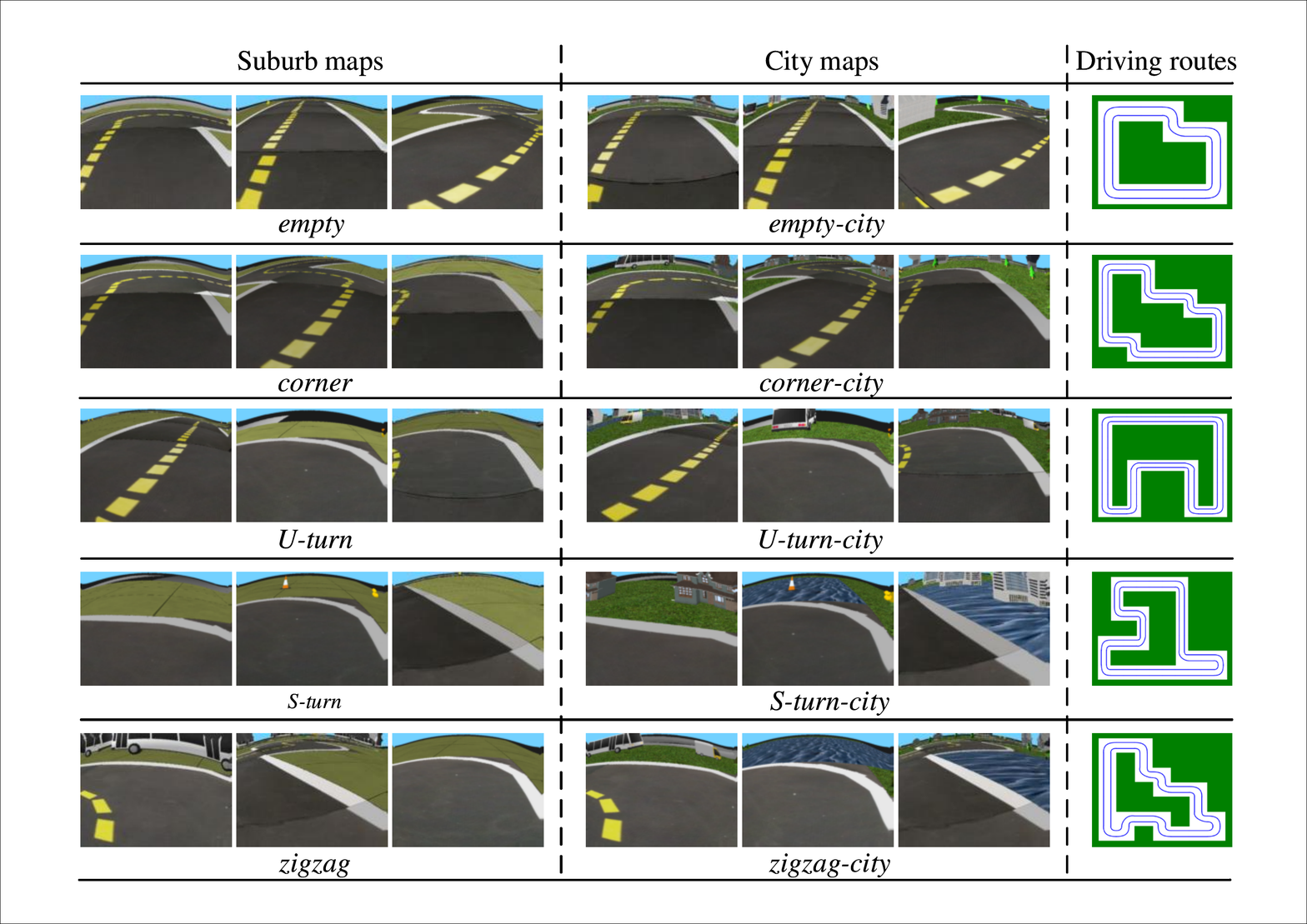}
  \end{center}
  \captionsetup{font={small}}
  \caption{Illustration of the background and driving routes of ten maps.}
  \label{fig:a1 maps}
 \end{figure}

 To evaluate our SSINet on Lane-following task based on Duckietown environment, we use ten maps which are mainly different from each other in background and driving route. These maps include two official maps (\emph{empty} and \emph{zigzag}) and eight customized maps (\emph{empty-city}, \emph{corner}, \emph{corner-city}, \emph{U-turn}, \emph{U-turn-city}, \emph{S-turn}, \emph{S-turn-city} and \emph{zigzag-city}). As shown in Fig. \ref{fig:a1 maps}, there are five different driving routes in total, each driving route corresponds to a map pair consisting of a suburb version and a city version, such as \emph{empty} and \emph{empty-city}. While the suburb maps' background only includes the grassland and a few ducks, the background of city maps is rich and includes a variety of objects, such as tree, water, grass, cone, bus, truck, house, building and duckie.

 Note that among five driving routes in the last column, the \emph{empty} is the easiest while the \emph{zigzag} is the most difficult. Specifically, the \emph{empty} has continuous left and right turns. The \emph{corner} has continuous multiple turns. The \emph{U-turn} has continuous multiple U-turns. The \emph{S-turn} has difficult S-turn and 180-degree turn. The \emph{zigzag} integrates continuous multiple turns, S-turn and 180-degree turn together.

%------------------------------------------------------------------------ 
 \begin{table}[t]
  \centering
  \caption{Hyperparameters used for PPO, SAC and TD3 in the first stage.}
  \label{tab:first stage}
  \begin{tabular}{l|l}
    \hline
    Hyperparameters & Value \\
    \hline
    \emph{Shared} \\
        \qquad optimizer & Adam \\
        \qquad start time steps & 1 ${\times 10^4}$ \\
        \qquad discount factor & 0.99 \\
        \qquad total frames & 500000 \\
        \qquad target network update rate & 0.005 \\
    \hline
    \emph{PPO} \\
        \qquad value loss coefficient & 0.5 \\
        \qquad entropy loss coefficient & 0.01 \\
        \qquad GAE coefficient & 0.95 \\
        \qquad ratio clip & 0.1 \\
        \qquad maximum gradient norm & 0.5 \\
        \qquad batch size & 256 \\
        \qquad learning rate & 0.00025 \\
    \hline
    \emph{SAC} \\
        \qquad batch size & 256 \\
        \qquad replay buffer size & 10000 \\
        \qquad temperature parameter & 0.2 \\
        \qquad learning rate for actor & 0.0003 \\
        \qquad learning rate for critic & 0.0003 \\
    \hline
    \emph{TD3} \\
        \qquad batch size & 32 \\
        \qquad explore noise & 0.1 \\
        \qquad policy noise & 0.2 \\
        \qquad noise clip & 0.5 \\
        \qquad delayed policy updates & 2 \\
        \qquad replay buffer size & 10000 \\
        \qquad learning rate for actor & 0.0001 \\
        \qquad learning rate for critic & 0.001 \\
    \hline
  \end{tabular}
 \end{table}
\section{Implementation Details}\label{app:details}
 As described in Section 4.3, the whole training procedure includes two stages. In the first stage, the \emph{feature extractor} and \emph{action predictor} are jointly pretrained with RL (here PPO, SAC or TD3), then the resulting expert policy is used to generate $M=50000$ state-action pairs. Main hyperparameters for pretraining the actor network are listed in Table \ref{tab:first stage}. In the second stage, the \emph{mask decoder} is trained with the state-action pairs generated in the first stage. Main hyperparameters for training the \emph{mask decoder} are listed in Table \ref{tab:second stage}.
 
%------------------------------------------------------------------------ 
 \begin{table}[t]
  \centering
  \caption{Hyperparameters used for training the \emph{mask decoder} in the second stage.}
  \label{tab:second stage}
  \begin{tabular}{l|l}
    \hline
    Hyperparameters & Value \\
    \hline
        \qquad optimizer & SGD \\
        \qquad epoch & 30 \\
        \qquad batch size & 32 \\
        \qquad learning rate & 0.01 \\
        \qquad momentum & 0.9 \\
        \qquad weight decay & 0.0005 \\
        \qquad regularization scale $\alpha$ & 0.04 \\
    \hline
  \end{tabular}
 \end{table}
\section{Additional Results of Section 5.2}\label{app:additional results}
\begin{figure}[h]
  \setlength{\abovecaptionskip}{-0.1cm}
  \setlength{\belowcaptionskip}{-0.3cm}
  \begin{center}
   \includegraphics*[width=0.95\linewidth]{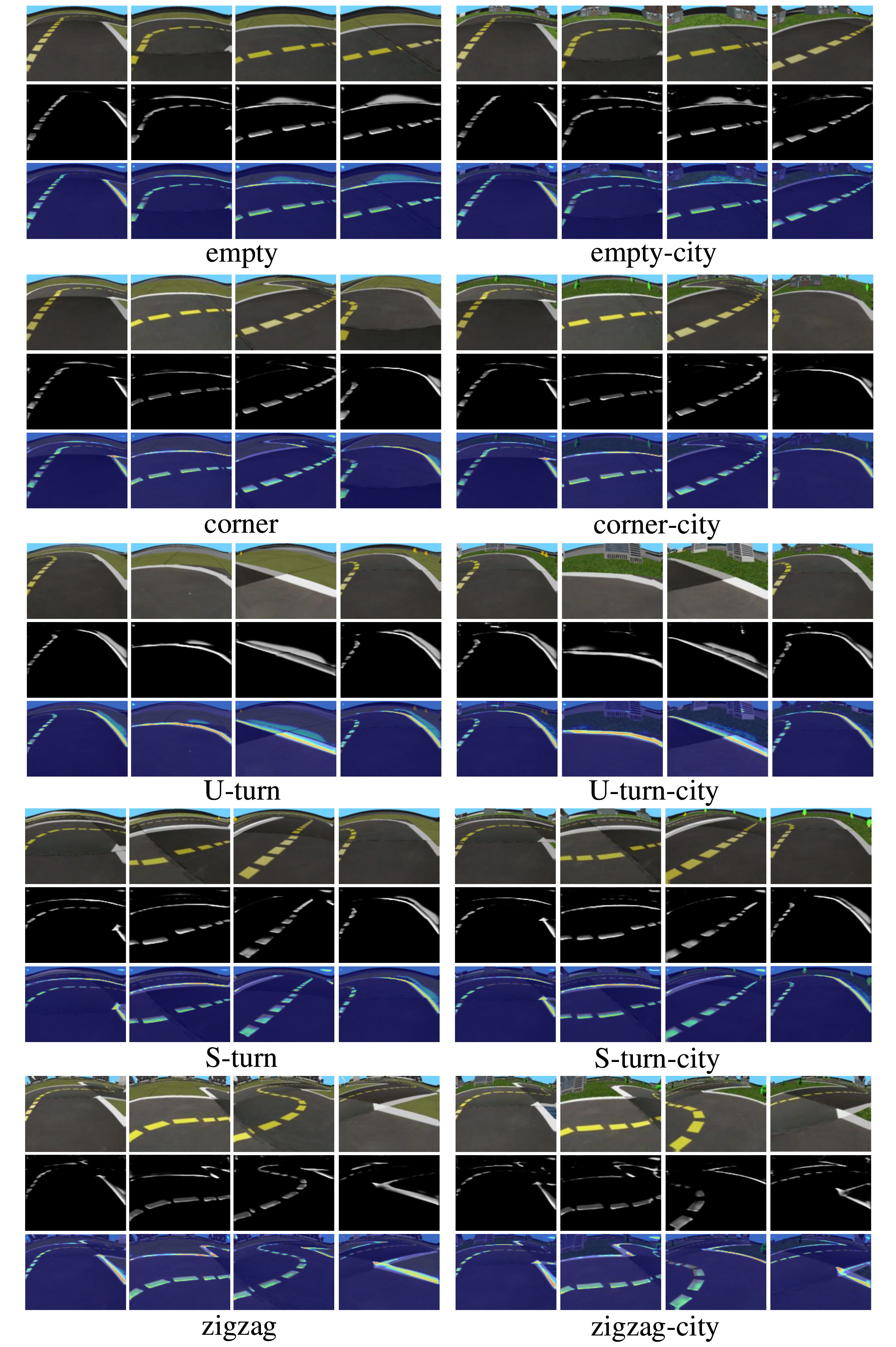}
  \end{center}
  \captionsetup{font={small}}
  \caption{Visualization of the performance on all maps. For each map, three rows represent a sequence of states, attention masks and masked states respectively (time goes from left to right).}
  \label{fig:b1 10maps}
 \end{figure}

 Fig. \ref{fig:b1 10maps} show the performance of our method on all maps. We observe that our method consistently produces high-resolution and sharp attention masks to highlight task-relevant information that constitutes most evidence for the predicted actions on all maps.
\vspace{-0.5em}

%use section* for acknowledgment
%\ifCLASSOPTIONcompsoc
%  % The Computer Society usually uses the plural form
%\else
%  % regular IEEE prefers the singular form
%\fi

\section*{Acknowledgments}
 This work is supported in part by the National Science and Technology Major Project of the Ministry of Science and Technology of China under Grants 2018AAA0100701 and 2018YFB1702903, the National Natural Science Foundation of China under Grants 61906106, 61936009 and 62022048, the Institute for Guo Qiang of Tsinghua University and Beijing Academy of Artificial Intelligence. We would like to thank the reviewers for their valuable comments.

\ifCLASSOPTIONcaptionsoff
  \newpage
\fi



\bibliographystyle{IEEEtran}

\bibliography{reference}

\begin{thebibliography}{10}
\providecommand{\url}[1]{#1}
\csname url@samestyle\endcsname
\providecommand{\newblock}{\relax}
\providecommand{\bibinfo}[2]{#2}
\providecommand{\BIBentrySTDinterwordspacing}{\spaceskip=0pt\relax}
\providecommand{\BIBentryALTinterwordstretchfactor}{4}
\providecommand{\BIBentryALTinterwordspacing}{\spaceskip=\fontdimen2\font plus
\BIBentryALTinterwordstretchfactor\fontdimen3\font minus
  \fontdimen4\font\relax}
\providecommand{\BIBforeignlanguage}[2]{{%
\expandafter\ifx\csname l@#1\endcsname\relax
\typeout{** WARNING: IEEEtran.bst: No hyphenation pattern has been}%
\typeout{** loaded for the language `#1'. Using the pattern for}%
\typeout{** the default language instead.}%
\else
\language=\csname l@#1\endcsname
\fi
#2}}
\providecommand{\BIBdecl}{\relax}
\BIBdecl

\bibitem{mnih2015human}
V.~Mnih, K.~Kavukcuoglu, D.~Silver, A.~A. Rusu, J.~Veness, M.~G. Bellemare,
  A.~Graves, M.~Riedmiller, A.~K. Fidjeland, G.~Ostrovski \emph{et~al.},
  ``Human-level control through deep reinforcement learning,'' \emph{Nature},
  vol. 518, no. 7540, p. 529, 2015.

\bibitem{silver2016mastering}
D.~Silver, A.~Huang, C.~J. Maddison, A.~Guez, L.~Sifre, G.~Van Den~Driessche,
  J.~Schrittwieser, I.~Antonoglou, V.~Panneershelvam, M.~Lanctot \emph{et~al.},
  ``Mastering the game of go with deep neural networks and tree search,''
  \emph{Nature}, vol. 529, no. 7587, p. 484, 2016.

\bibitem{shi2019regularized}
W.~Shi, S.~Song, H.~Wu, Y.-C. Hsu, C.~Wu, and G.~Huang, ``Regularized anderson
  acceleration for off-policy deep reinforcement learning,'' in \emph{Advances
  in Neural Information Processing Systems}, 2019, pp. 10\,231--10\,241.

\bibitem{mirowski2016learning}
P.~Mirowski, R.~Pascanu, F.~Viola, H.~Soyer, A.~J. Ballard, A.~Banino,
  M.~Denil, R.~Goroshin, L.~Sifre, K.~Kavukcuoglu \emph{et~al.}, ``Learning to
  navigate in complex environments,'' in \emph{Proceedings of the International
  Conference on Learning Representations}, 2017.

\bibitem{shi2018multi}
W.~Shi, S.~Song, C.~Wu, and C.~P. Chen, ``Multi pseudo q-learning-based
  deterministic policy gradient for tracking control of autonomous underwater
  vehicles,'' \emph{IEEE Transactions on Neural Networks and Learning Systems},
  vol.~30, no.~12, pp. 3534--3546, 2018.

\bibitem{shi2018high}
W.~Shi, S.~Song, and C.~Wu, ``High-level tracking of autonomous underwater
  vehicles based on pseudo averaged q-learning,'' in \emph{2018 IEEE
  International Conference on Systems, Man, and Cybernetics (SMC)}.\hskip 1em
  plus 0.5em minus 0.4em\relax IEEE, 2018, pp. 4138--4143.

\bibitem{cao2019interpretable}
Q.~Cao, X.~Liang, B.~Li, and L.~Lin, ``Interpretable visual question answering
  by reasoning on dependency trees,'' \emph{IEEE Transactions on Pattern
  Analysis and Machine Intelligence}, 2019.

\bibitem{monfort2019moments}
M.~Monfort, A.~Andonian, B.~Zhou, K.~Ramakrishnan, S.~A. Bargal, T.~Yan,
  L.~Brown, Q.~Fan, D.~Gutfreund, C.~Vondrick \emph{et~al.}, ``Moments in time
  dataset: one million videos for event understanding,'' \emph{IEEE
  Transactions on Pattern Analysis and Machine Intelligence}, vol.~42, no.~2,
  pp. 502--508, 2019.

\bibitem{liu2019tabby}
H.~Liu, R.~Wang, S.~Shan, and X.~Chen, ``What is tabby? interpretable model
  decisions by learning attribute-based classification criteria,'' \emph{IEEE
  Transactions on Pattern Analysis and Machine Intelligence}, 2019.

\bibitem{guidotti2019survey}
R.~Guidotti, A.~Monreale, S.~Ruggieri, F.~Turini, F.~Giannotti, and
  D.~Pedreschi, ``A survey of methods for explaining black box models,''
  \emph{ACM Computing Surveys (CSUR)}, vol.~51, no.~5, p.~93, 2019.

\bibitem{ribeiro2016should}
M.~T. Ribeiro, S.~Singh, and C.~Guestrin, ``"why should i trust you?"
  explaining the predictions of any classifier,'' in \emph{Proceedings of the
  22nd ACM SIGKDD International Conference on Knowledge Discovery and Data
  Mining}, 2016, pp. 1135--1144.

\bibitem{binder2016layer}
A.~Binder, G.~Montavon, S.~Lapuschkin, K.-R. M{\"u}ller, and W.~Samek,
  ``Layer-wise relevance propagation for neural networks with local
  renormalization layers,'' in \emph{International Conference on Artificial
  Neural Networks}.\hskip 1em plus 0.5em minus 0.4em\relax Springer, 2016, pp.
  63--71.

\bibitem{shrikumar2016not}
A.~Shrikumar, P.~Greenside, A.~Shcherbina, and A.~Kundaje, ``Not just a black
  box: Learning important features through propagating activation
  differences,'' \emph{arXiv preprint arXiv:1605.01713}, 2016.

\bibitem{selvaraju2017grad}
R.~R. Selvaraju, M.~Cogswell, A.~Das, R.~Vedantam, D.~Parikh, and D.~Batra,
  ``Grad-cam: Visual explanations from deep networks via gradient-based
  localization,'' in \emph{Proceedings of the IEEE International Conference on
  Computer Vision}, 2017, pp. 618--626.

\bibitem{lundberg2017unified}
S.~M. Lundberg and S.-I. Lee, ``A unified approach to interpreting model
  predictions,'' in \emph{Advances in Neural Information Processing Systems},
  2017, pp. 4765--4774.

\bibitem{bau2017network}
D.~Bau, B.~Zhou, A.~Khosla, A.~Oliva, and A.~Torralba, ``Network dissection:
  Quantifying interpretability of deep visual representations,'' in
  \emph{Proceedings of the IEEE Conference on Computer Vision and Pattern
  Recognition}, 2017, pp. 6541--6549.

\bibitem{zhou2018interpreting}
B.~Zhou, D.~Bau, A.~Oliva, and A.~Torralba, ``Interpreting deep visual
  representations via network dissection,'' \emph{IEEE Transactions on Pattern
  Analysis and Machine Intelligence}, vol.~41, no.~9, pp. 2131--2145, 2018.

\bibitem{zahavy2016graying}
T.~Zahavy, N.~Ben-Zrihem, and S.~Mannor, ``Graying the black box: Understanding
  dqns,'' in \emph{International Conference on Machine Learning}, 2016, pp.
  1899--1908.

\bibitem{annasamy2019towards}
R.~M. Annasamy and K.~Sycara, ``Towards better interpretability in deep
  q-networks,'' in \emph{Proceedings of the AAAI Conference on Artificial
  Intelligence}, vol.~33, 2019, pp. 4561--4569.

\bibitem{wang2016dueling}
Z.~Wang, T.~Schaul, M.~Hessel, H.~Van~Hasselt, M.~Lanctot, and N.~De~Freitas,
  ``Dueling network architectures for deep reinforcement learning,'' in
  \emph{International Conference on Machine Learning}, vol.~48, 2016, pp.
  1995--2003.

\bibitem{greydanus2018visualizing}
S.~Greydanus, A.~Koul, J.~Dodge, and A.~Fern, ``Visualizing and understanding
  atari agents,'' in \emph{International Conference on Machine Learning}, 2018.

\bibitem{manchin2019reinforcement}
A.~Manchin, E.~Abbasnejad, and A.~v.~d. Hengel, ``Reinforcement learning with
  attention that works: A self-supervised approach,'' \emph{arXiv preprint
  arXiv:1904.03367}, 2019.

\bibitem{mott2019towards}
A.~Mott, D.~Zoran, M.~Chrzanowski, D.~Wierstra, and D.~Jimenez~Rezende,
  ``Towards interpretable reinforcement learning using attention augmented
  agents,'' in \emph{Advances in Neural Information Processing Systems}, 2019,
  pp. 12\,350--12\,359.

\bibitem{zhang2018agil}
R.~Zhang, Z.~Liu, L.~Zhang, J.~A. Whritner, K.~S. Muller, M.~M. Hayhoe, and
  D.~H. Ballard, ``Agil: Learning attention from human for visuomotor tasks,''
  in \emph{Proceedings of the European Conference on Computer Vision}, 2018,
  pp. 663--679.

\bibitem{zhang2019atari}
R.~Zhang, Z.~Liu, L.~Guan, L.~Zhang, M.~M. Hayhoe, and D.~H. Ballard,
  ``Atari-head: Atari human eye-tracking and demonstration dataset,''
  \emph{arXiv preprint arXiv:1903.06754}, 2019.

\bibitem{bellemare2013arcade}
M.~G. Bellemare, Y.~Naddaf, J.~Veness, and M.~Bowling, ``The arcade learning
  environment: An evaluation platform for general agents,'' \emph{Journal of
  Artificial Intelligence Research}, vol.~47, pp. 253--279, 2013.

\bibitem{gym_duckietown}
M.~Chevalier-Boisvert, F.~Golemo, Y.~Cao, B.~Mehta, and L.~Paull, ``Duckietown
  environments for openai gym,''
  \url{https://github.com/duckietown/gym-duckietown}, 2018.

\bibitem{dodson2011natural}
T.~Dodson, N.~Mattei, and J.~Goldsmith, ``A natural language argumentation
  interface for explanation generation in markov decision processes,'' in
  \emph{International Conference on Algorithmic Decision Theory}.\hskip 1em
  plus 0.5em minus 0.4em\relax Springer, 2011, pp. 42--55.

\bibitem{elizalde2008policy}
F.~Elizalde, L.~E. Sucar, M.~Luque, J.~Diez, and A.~Reyes, ``Policy explanation
  in factored markov decision processes,'' in \emph{Proceedings of the 4th
  European Workshop on Probabilistic Graphical Models (PGM 2008)}, 2008, pp.
  97--104.

\bibitem{hayes2017improving}
B.~Hayes and J.~A. Shah, ``Improving robot controller transparency through
  autonomous policy explanation,'' in \emph{12th ACM/IEEE International
  Conference on Human-Robot Interaction}, 2017, pp. 303--312.

\bibitem{gupta2015policy}
U.~D. Gupta, E.~Talvitie, and M.~Bowling, ``Policy tree: Adaptive
  representation for policy gradient,'' in \emph{Twenty-Ninth AAAI Conference
  on Artificial Intelligence}, 2015.

\bibitem{bastani2018verifiable}
O.~Bastani, Y.~Pu, and A.~Solar-Lezama, ``Verifiable reinforcement learning via
  policy extraction,'' in \emph{Advances in Neural Information Processing
  Systems}, 2018, pp. 2494--2504.

\bibitem{roth2019conservative}
A.~M. Roth, N.~Topin, P.~Jamshidi, and M.~Veloso, ``Conservative q-improvement:
  Reinforcement learning for an interpretable decision-tree policy,''
  \emph{arXiv preprint arXiv:1907.01180}, 2019.

\bibitem{waa2018contrastive}
J.~Waa, J.~v. Diggelen, K.~Bosch, and M.~Neerincx, ``Contrastive explanations
  for reinforcement learning in terms of expected consequences,'' in
  \emph{Proceedings of the Workshop on Explainable AI on the IJCAI}, 2018.

\bibitem{madumal2019explainable}
P.~Madumal, T.~Miller, L.~Sonenberg, and F.~Vetere, ``Explainable reinforcement
  learning through a causal lens,'' \emph{arXiv preprint arXiv:1905.10958},
  2019.

\bibitem{maaten2008visualizing}
L.~v.~d. Maaten and G.~Hinton, ``Visualizing data using t-sne,'' \emph{Journal
  of Machine Learning Research}, vol.~9, no. Nov, pp. 2579--2605, 2008.

\bibitem{engel2001learning}
Y.~Engel and S.~Mannor, ``Learning embedded maps of markov processes,'' in
  \emph{International Conference on Machine Learning}, 2001.

\bibitem{simonyan2014deep}
K.~Simonyan, A.~Vedaldi, and A.~Zisserman, ``Deep inside convolutional
  networks: Visualising image classification models and saliency maps,'' in
  \emph{Workshop Proceedings of the International Conference on Learning
  Representations}, 2014.

\bibitem{zhang2018top}
J.~Zhang, S.~A. Bargal, Z.~Lin, J.~Brandt, X.~Shen, and S.~Sclaroff, ``Top-down
  neural attention by excitation backprop,'' \emph{International Journal of
  Computer Vision}, vol. 126, no.~10, pp. 1084--1102, 2018.

\bibitem{smilkov2017smoothgrad}
D.~Smilkov, N.~Thorat, B.~Kim, F.~Vi{\'e}gas, and M.~Wattenberg, ``Smoothgrad:
  removing noise by adding noise,'' \emph{arXiv preprint arXiv:1706.03825},
  2017.

\bibitem{zintgraf2017visualizing}
L.~M. Zintgraf, T.~S. Cohen, T.~Adel, and M.~Welling, ``Visualizing deep neural
  network decisions: Prediction difference analysis,'' \emph{arXiv preprint
  arXiv:1702.04595}, 2017.

\bibitem{fong2017interpretable}
R.~C. Fong and A.~Vedaldi, ``Interpretable explanations of black boxes by
  meaningful perturbation,'' in \emph{Proceedings of the IEEE International
  Conference on Computer Vision}, 2017, pp. 3429--3437.

\bibitem{zeiler2014visualizing}
M.~D. Zeiler and R.~Fergus, ``Visualizing and understanding convolutional
  networks,'' in \emph{Proceedings of the European Conference on Computer
  Vision}.\hskip 1em plus 0.5em minus 0.4em\relax Springer, 2014, pp. 818--833.

\bibitem{petsiuk2018rise}
V.~Petsiuk, A.~Das, and K.~Saenko, ``Rise: Randomized input sampling for
  explanation of black-box models,'' in \emph{British Machine Vision
  Conference}, 2018.

\bibitem{shapley1953value}
L.~S. Shapley, ``A value for n-person games,'' \emph{Contributions to the
  Theory of Games}, vol.~2, no.~28, pp. 307--317, 1953.

\bibitem{ancona2019explaining}
M.~Ancona, C.~{\"O}ztireli, and M.~Gross, ``Explaining deep neural networks
  with a polynomial time algorithm for shapley values approximation,'' in
  \emph{International Conference on Machine Learning}, 2019.

\bibitem{fong2019understanding}
R.~Fong, M.~Patrick, and A.~Vedaldi, ``Understanding deep networks via extremal
  perturbations and smooth masks,'' in \emph{Proceedings of the IEEE
  International Conference on Computer Vision}, 2019, pp. 2950--2958.

\bibitem{dabkowski2017real}
P.~Dabkowski and Y.~Gal, ``Real time image saliency for black box
  classifiers,'' in \emph{Advances in Neural Information Processing Systems},
  2017, pp. 6967--6976.

\bibitem{wang2020paying}
W.~Wang, J.~Shen, X.~Lu, S.~C. Hoi, and H.~Ling, ``Paying attention to video
  object pattern understanding,'' \emph{IEEE Transactions on Pattern Analysis
  and Machine Intelligence}, 2020.

\bibitem{sorokin2015deep}
I.~Sorokin, A.~Seleznev, M.~Pavlov, A.~Fedorov, and A.~Ignateva, ``Deep
  attention recurrent q-network,'' \emph{arXiv preprint arXiv:1512.01693},
  2015.

\bibitem{mousavi2016learning}
S.~Mousavi, M.~Schukat, E.~Howley, A.~Borji, and N.~Mozayani, ``Learning to
  predict where to look in interactive environments using deep recurrent
  q-learning,'' \emph{arXiv preprint arXiv:1612.05753}, 2016.

\bibitem{yang2018learn}
Z.~Yang, S.~Bai, L.~Zhang, and P.~H. Torr, ``Learn to interpret atari agents,''
  \emph{arXiv preprint arXiv:1812.11276}, 2018.

\bibitem{nikulin2019free}
D.~Nikulin, A.~Ianina, V.~Aliev, and S.~Nikolenko, ``Free-lunch saliency via
  attention in atari agents,'' \emph{arXiv preprint arXiv:1908.02511}, 2019.

\bibitem{choi2017multi}
J.~Choi, B.-J. Lee, and B.-T. Zhang, ``Multi-focus attention network for
  efficient deep reinforcement learning,'' in \emph{Workshops at the
  Thirty-First AAAI Conference on Artificial Intelligence}, 2017.

\bibitem{schulman2017proximal}
J.~Schulman, F.~Wolski, P.~Dhariwal, A.~Radford, and O.~Klimov, ``Proximal
  policy optimization algorithms,'' \emph{arXiv preprint arXiv:1707.06347},
  2017.

\bibitem{haarnoja2018soft}
T.~Haarnoja, A.~Zhou, P.~Abbeel, and S.~Levine, ``Soft actor-critic: Off-policy
  maximum entropy deep reinforcement learning with a stochastic actor,'' in
  \emph{International Conference on Machine Learning}, 2018.

\bibitem{fujimoto2018addressing}
S.~Fujimoto, H.~van Hoof, and D.~Meger, ``Addressing function approximation
  error in actor-critic methods,'' in \emph{International Conference on Machine
  Learning}, 2018.

\bibitem{shi2019soft}
W.~Shi, S.~Song, and C.~Wu, ``Soft policy gradient method for maximum entropy
  deep reinforcement learning,'' in \emph{Proceedings of the 28th International
  Joint Conference on Artificial Intelligence}, 2019, pp. 3425--3431.

\bibitem{van2016deep}
H.~Van~Hasselt, A.~Guez, and D.~Silver, ``Deep reinforcement learning with
  double q-learning,'' in \emph{Thirtieth AAAI Conference on Artificial
  Intelligence}, 2016.

\bibitem{lillicrap2015continuous}
T.~P. Lillicrap, J.~J. Hunt, A.~Pritzel, N.~Heess, T.~Erez, Y.~Tassa,
  D.~Silver, and D.~Wierstra, ``Continuous control with deep reinforcement
  learning,'' in \emph{International Conference on Learning Representations},
  2016.

\bibitem{ronneberger2015u}
O.~Ronneberger, P.~Fischer, and T.~Brox, ``U-net: Convolutional networks for
  biomedical image segmentation,'' in \emph{International Conference on Medical
  Image Computing and Computer-Assisted Intervention}.\hskip 1em plus 0.5em
  minus 0.4em\relax Springer, 2015, pp. 234--241.

\bibitem{chen2017rethinking}
L.-C. Chen, G.~Papandreou, F.~Schroff, and H.~Adam, ``Rethinking atrous
  convolution for semantic image segmentation,'' \emph{arXiv preprint
  arXiv:1706.05587}, 2017.

\bibitem{mayer2016large}
N.~Mayer, E.~Ilg, P.~Hausser, P.~Fischer, D.~Cremers, A.~Dosovitskiy, and
  T.~Brox, ``A large dataset to train convolutional networks for disparity,
  optical flow, and scene flow estimation,'' in \emph{Proceedings of the IEEE
  Conference on Computer Vision and Pattern Recognition}, 2016, pp. 4040--4048.

\bibitem{openai2016gym}
G.~Brockman, V.~Cheung, L.~Pettersson, J.~Schneider, J.~Schulman, J.~Tang, and
  W.~Zaremba, ``Openai gym,'' 2016.

\bibitem{land2009vision}
M.~F. Land, ``Vision, eye movements, and natural behavior,'' \emph{Visual
  Neuroscience}, vol.~26, no.~1, pp. 51--62, 2009.

\bibitem{lin2017refinenet}
G.~Lin, A.~Milan, C.~Shen, and I.~Reid, ``Refinenet: Multi-path refinement
  networks for high-resolution semantic segmentation,'' in \emph{Proceedings of
  the IEEE Conference on Computer Vision and Pattern Recognition}, 2017, pp.
  1925--1934.

\bibitem{jegou2017one}
S.~J{\'e}gou, M.~Drozdzal, D.~Vazquez, A.~Romero, and Y.~Bengio, ``The one
  hundred layers tiramisu: Fully convolutional densenets for semantic
  segmentation,'' in \emph{Proceedings of the IEEE Conference on Computer
  Vision and Pattern Recognition Workshops}, 2017, pp. 11--19.

\end{thebibliography}
\end{document}